\documentclass[10pt,journal,compsoc]{IEEEtran}
\ifCLASSOPTIONcompsoc
  \usepackage[nocompress]{cite}
\else
  \usepackage{cite}
\fi
\usepackage[utf8]{inputenc}

\usepackage{microtype}
\usepackage{algorithm,algcompatible,amsmath}
\usepackage{graphicx}
\usepackage{url}
\usepackage{multirow}
\usepackage{comment}
\usepackage{amssymb}
\usepackage{pifont}
\usepackage{mathtools}
\usepackage{xcolor}
\usepackage{enumitem}
\usepackage{marvosym}
\usepackage{textcomp}
\usepackage{array}
\usepackage{booktabs}
\usepackage{multicol}
\usepackage{flushend}
\usepackage{amsfonts}
\usepackage{amsmath}
\usepackage{xcolor}
\usepackage{calc}
\usepackage{amssymb}
\usepackage{diagbox}
\usepackage{algorithm}  
\usepackage{algorithmicx}  
\usepackage{algpseudocode} 
\usepackage{ulem}
\usepackage{color}
\usepackage{array}
\makeatletter
\newcommand{\thickhline}{%
    \noalign {\ifnum 0=`}\fi \hrule height 1pt
    \futurelet \reserved@a \@xhline
}

\newcommand{\squishend}{
\end{list} }
\newcommand*\circled[1]{\kern-2.5em%
  \put(0,4){\color{white}\circle*{18}}\put(0,4){\circle{10}}%
  \put(-3,0){\color{black}\bfseries#1}~~}
\usepackage{tikz} 

\usepackage{enumitem}

\newcommand{\squishlist}{
\begin{list}{$\bullet$}
{ \usecounter{Lcount}
\setlength{\itemsep}{0pt}
\setlength{\parsep}{0pt}
\setlength{\topsep}{0pt}
\setlength{\partopsep}{0pt}
\setlength{\leftmargin}{2em}
\setlength{\labelwidth}{1.5em}
\setlength{\labelsep}{0.5em} } }
\ifCLASSINFOpdf
\else
\fi
\newcommand{\modelname}{\texttt{GIRL}}

\newcommand{\gradient}{Gradient Imitation Reinforcement Learning}

\begin{document}
%
\title{Gradient Imitation Reinforcement Learning for General Low-Resource Information Extraction}
%
%
%
%

\author{Xuming~Hu, Shiao~Meng, Chenwei~Zhang, Xiangli~Yang, Lijie~Wen,\\ Irwin~King,~\IEEEmembership{Fellow,~IEEE}, and Philip~S.~Yu,~\IEEEmembership{Life Fellow,~IEEE}
\IEEEcompsocitemizethanks{\IEEEcompsocthanksitem X. Hu, S. Meng and L. Wen are with the School of Software, Tsinghua University, Beijing, China\protect\\
E-mail: \{hxm19, msa21\}@mails.tsinghua.edu.cn, wenlj@tsinghua.edu.cn
\IEEEcompsocthanksitem C. Zhang is with the Amazon, Seattle, USA\protect\\
E-mail: cwzhang910@gmail.com
\IEEEcompsocthanksitem X. Yang is with the School of Computer Science and Engineering, University of Electronic Science and Technology of China, Chengdu, China\protect\\
E-mail: xlyang@std.uestc.edu.cn
\IEEEcompsocthanksitem I. King is with the Department of Computer Science and Engineering, The Chinese University of Hong Kong, Hong Kong, China\protect\\
E-mail: king@cse.cuhk.edu.hk
\IEEEcompsocthanksitem P. S. Yu is with the Department of Computer Science, University of Illinois at Chicago, Chicago, USA\protect\\
E-mail: psyu@uic.edu
\IEEEcompsocthanksitem Corresponding authors: Chenwei Zhang and Lijie Wen.}
\thanks{Manuscript received xx xx, xx; revised xx xx, xx.}}

%
%

\markboth{Journal of \LaTeX\ Class Files,~Vol.~14, No.~8, August~2015}%
{Shell \MakeLowercase{\textit{et al.}}: Bare Demo of IEEEtran.cls for Computer Society Journals}
%



\IEEEtitleabstractindextext{%
\begin{abstract}
Information Extraction (IE) aims to extract structured information from heterogeneous sources. IE from natural language texts include sub-tasks such as Named Entity Recognition (NER), Relation Extraction (RE), and Event Extraction (EE).
Most IE systems require comprehensive understandings of sentence structure, implied semantics, and domain knowledge to perform well; thus, IE tasks always need adequate external resources and annotations. However, it takes time and effort to obtain more human annotations. Low-Resource Information Extraction (LRIE) strives to use unsupervised data, reducing the required resources and human annotation. In practice, existing systems either utilize self-training schemes to generate pseudo labels that will cause the gradual drift problem, or leverage consistency regularization methods which inevitably possess confirmation bias. To alleviate confirmation bias due to the lack of feedback loops in existing LRIE learning paradigms, we develop a {\gradient} ({\modelname}) method to encourage pseudo-labeled data to imitate the gradient descent direction on labeled data, which can force pseudo-labeled data to achieve better optimization capabilities similar to labeled data. Based on how well the pseudo-labeled data imitates the instructive gradient descent direction obtained from labeled data, we design a reward to quantify the imitation process and bootstrap the optimization capability of pseudo-labeled data through trial and error. In addition to learning paradigms, {\modelname} is not limited to specific sub-tasks, and we leverage {\modelname} to solve all IE sub-tasks (named entity recognition, relation extraction, and event extraction) in low-resource settings (semi-supervised IE and few-shot IE).
Experimental results on seven public datasets across three IE sub-tasks demonstrate the effectiveness of {\modelname} on low-resource information extraction when comparing with strong baselines. 
\end{abstract}

\begin{IEEEkeywords}
Information Extraction, Low-Resource, General Framework, Gradient Optimization, Reinforcement Learning.
\end{IEEEkeywords}}

\maketitle

\IEEEdisplaynontitleabstractindextext

%
\IEEEpeerreviewmaketitle

\IEEEraisesectionheading{
\section{Introduction}\label{sec:introduction}}

\IEEEPARstart{L}{arge} amounts of human knowledge has been carried out in natural language. Although this knowledge is crucial for many applications, such as question answering \cite{allam2012question}, market analysis \cite{isinkaye2015recommendation}, and search engines \cite{madhu2011intelligent}. Most of this massive unstructured knowledge is inaccessible to computers and difficult to obtain by human experts. To organize and retrieve knowledge, computer systems rely on structured representations (such as databases).
Therefore, Information Extraction (IE) -- discovering structured information from massive heterogeneous sources -- is a crucial step to increase the accessibility and availability of human knowledge. More specifically, IE needs to transform the unstructured knowledge into relational triplets, which are composed of a set of arguments and a phrase signifying a semantic relationship between them. To deal with the complexity and ambiguity of human language, IE tasks need to fully understand the natural language, including words, syntax, semantics, etc.

We set our focus on three typical types of sub-tasks in IE: Named Entity Recognition (NER), Relation Extraction (RE), and Event Extraction (EE). The NER task aims to infer a label for each token in the sentence to indicate whether it belongs to an entity and classify entities into predefined
\begin{figure}[t!]
    \centering
    \includegraphics[width=0.99\linewidth]{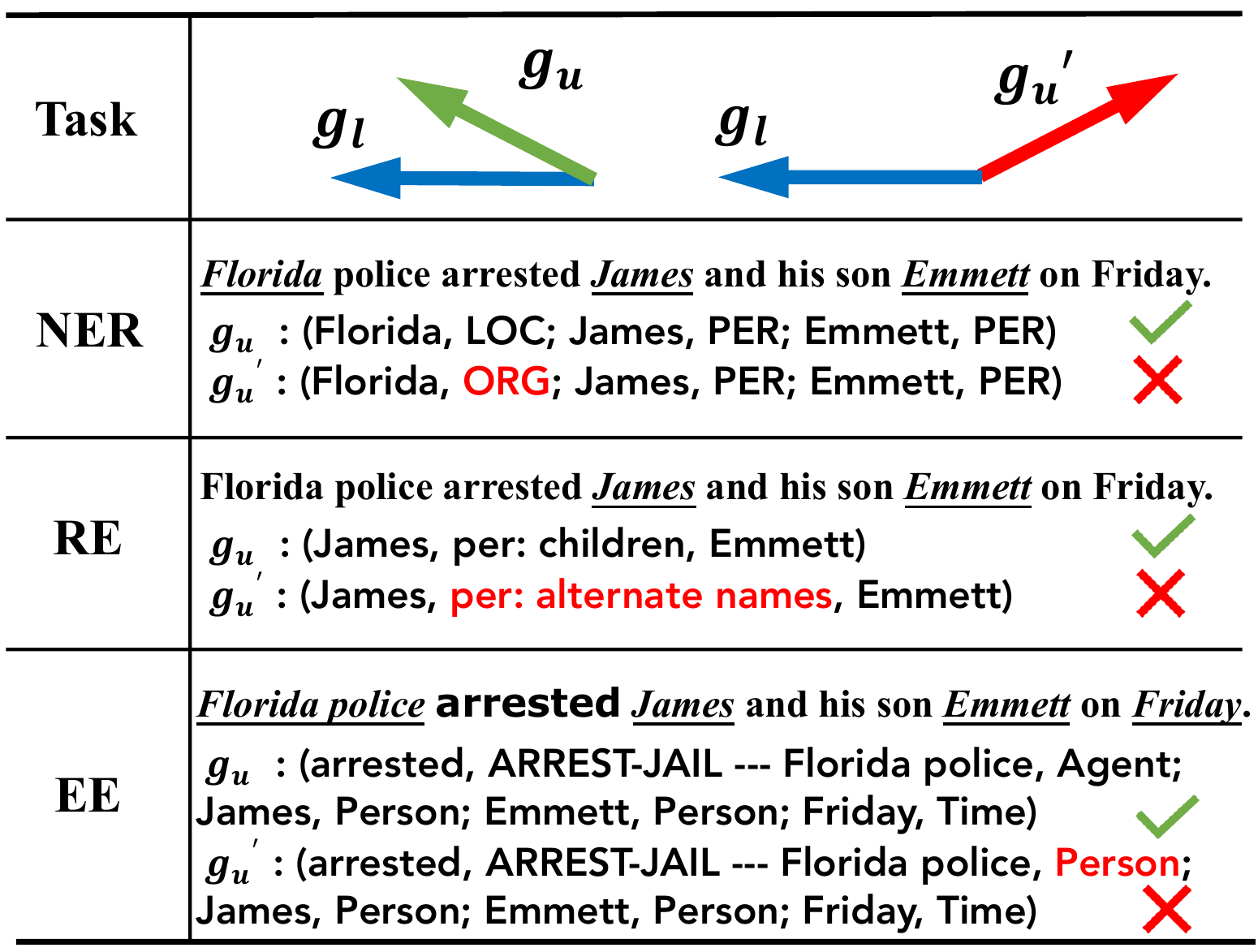}
    \caption{Gradient descent direction on labeled data (${g_{l}}$) and unlabeled data with correct or \textcolor{red}{incorrect} pseudo label (${g_{u}}$, ${g_{u}^{'}}$) for three Information Extraction sub-tasks.}
    \label{fig:introduction}
    \vspace{-1mm}
\end{figure}
types~\cite{sharma2022named}. As illustrated in Figure \ref{fig:introduction}, we can infer entities ``Florida, James, Emmett'' for the sentence ``Florida police arrested James and his son Emmett on Friday'' and label them with the corresponding entity types as ``Location (LOC), Person (PER), Person (PER)''.
After we obtain entities from the sentences, the RE task aims to discover the semantic relation that holds between two entities and transforms massive corpus into structured triplets (head entity, relation, tail entity)~\cite{wang2022deep}. For example, from ``Florida police arrested \textit{James}$_{head}$ and his son \textit{Emmett}$_{tail}$ on Friday", we can extract a relation ${\texttt{per:children}}$ indicating a parent-child relation between head and tail entities. EE task is a more complicated task that aims to convert a sentence into an event record, which usually consists of triggers and arguments~\cite{afyouni2022multi}. For example, in Figure~\ref{fig:introduction}, EE maps the sentence ``Florida police arrested James and his son Emmett on Friday'' into an event record: \{Trigger: arrested, Event Type: ARREST-JAIL --- Arg1: Florida Police, Arg1 Role: Agent, Arg2: James, Arg2 Role: Person, Arg3: Emmett, Arg3 Role: Person, Arg4: Friday, Arg4 Role: Time\}. 
Obviously, such structured information requires IE systems' comprehensive understandings of sentence structure, implied semantics, and domain knowledge; thus, IE tasks always need adequate human annotations. However, due to the explosion of knowledge and information, newly created entities, relations, and events come from different domains and in different forms. Therefore, it is difficult for the IE systems to obtain more human annotations. This motivates a Low-Resource Information Extraction (LRIE) task where human annotations are scarce~\cite{holzenberger2022asking,deng2022low,liu2022hierarchical}.

To handle the LRIE task, many efforts are devoted to improving the model generalization ability beyond learning directly from existing, limited human annotations. Distant supervision method is a classic way to solve the LRIE task, leveraging facts stored in external knowledge bases (KBs) to obtain annotated entities, relations, and triggers as supervision \cite{zhou2021self}. However, these methods all make a strong assumption that co-occurring entities convey KB entity types, relations, and argument roles regardless of specific contexts, which makes IE models generate structured information based on contextless rules and limits the generalization ability. To leverage unlabeled data, Rosenberg \textit{et al.} \cite{rosenberg2005semi} propose a self-training method, which aims to predict pseudo labels on unlabeled data and leverage pseudo-labeled data to iteratively improve the generalization capability of the IE models. However, during the training process, the self-training model suffers from the gradual drift problem \cite{curran2007minimising,zhang2016understanding,zhou2022active} caused by noisy pseudo-labeled data. 
Tarvainen and Valpola \cite{tarvainen2017mean} alleviate the noise in pseudo labels by adopting consistency regularizations based on the smoothness assumption which requires that perturbations to the data embedding will not affect the output predictions~\cite{oliver2018realistic,ke2019dual}.
However, when limited annotations are directly used during training, trained models inevitably possess confirmation bias towards~\cite{tarvainen2017mean}, if not overfits on, limited annotations, preventing LRIE models from further generalizing beyond the limited annotations.

To improve the generalization ability for LRIE models, we propose to use existing annotations as a guideline instead of having them directly involved in training, as well as introduce an explicit feedback loop when consuming annotations.
More specifically, we first encourage pseudo-labeled data to imitate labeled data on the gradient descent directions during the optimization process.
We illustrate this idea in Figure \ref{fig:introduction}, where ${g_{l}}$ represents the average gradient descent direction on labeled data. ${g_{u}}$ and ${g_{u}^{'}}$ represent the correct and incorrect pseudo labels on unlabeled data, which guides the gradient descent direction in a positive/negative fashion \cite{du2018adapting,sariyildiz2019gradient,yu2020gradient}. 
Based on how well the pseudo-labeled data mimics the instructive gradient descent direction obtained from limited labeled data, we then design a reward to quantify the behavior and aim to use the reward as an explicit feedback.
This learnable setting can be naturally formulated into a reinforcement learning framework, which aims to learn an imitation policy that maximizes the reward through trial and error. When comparing with methods where annotations are directly used in the traditional learning schema, this formulation also allows a feedback mechanism and thus increases generalization ability beyond limited annotations. We name our method as {\gradient} ({\modelname}) in this paper.

Furthermore, our model is not limited to specific sub-tasks. For general IE tasks, such as labeling pseudo entity types in NER, and labeling entity pairs with pseudo relation labels in RE, etc., {\modelname} could flexibly guide pseudo-labeled data to mimic the gradient descent direction obtained by limited labeled data.
To verify the generality and generalization ability of {\modelname}, we first develop the vanilla models of three IE sub-tasks (named entity recognition, relation extraction, and event extraction), and further leverage {\modelname} to solve all IE sub-tasks in low-resource settings (semi-supervised IE and few-shot IE). 
Our contributions are as follows:
\begin{itemize}
\item We propose the {\gradient} ({\modelname}) method for general low-resource IE tasks, which could alleviate the bias from training directly with limited annotation, and encourage the IE models to effectively generalize beyond limited annotations.
\item We demonstrate the effectiveness of {\modelname} in three major IE sub-tasks (named entity recognition, relation extraction, and event extraction) and experiment with both regular and low-resource settings (semi-supervised and few-shot IE).
\item We show that {\modelname} outperforms strong baselines on seven public datasets across three sub-tasks. Extensive experiments validate the effectiveness of {\modelname}.\footnote{We will open source data and code upon acceptance.}
\end{itemize}


\section{Related Work}\label{sec:related}
\subsection{Deep Low-resource Information Extraction Methods}
Information Extraction aims to extract structured information from heterogeneous sources. Recent literature leverages deep neural networks to encode the features of entities, relations, and events from sentences, and then classifies these features into pre-defined specific entity types, relation types, event types, and argument roles. These methods could gain decent performance when sufficient labeled data is available \cite{zhang2017position,zhou2021self}. However, it is often labor-intensive and requires expert knowledge to obtain large amounts of manual annotations.

Low-resource Information Extraction methods gained a lot of attention recently \cite{hu2020semi,hu2020selfore,hu2021gradient}, since these methods require fewer labeled data and deep neural networks could expand limited labeled information by exploiting information on unlabeled data to improve the model performance. In this paper, we divide deep low-resource methods into five groups \cite{yang2021survey}.

(1) Generative methods. To exploit the distribution of the training dataset and subsequently create new samples, generative models such as Variational Auto-Encoders (VAEs)~\cite{xu2017variational}, Generative Adversarial Networks (GANs)~\cite{goodfellow2014generative,zhu2018generative}, and their variations have evolved. On the basis of these structures, low-resource generative approaches have been researched and formalized~\cite{denton2016semi,dai2017good,wei2018improving}.

(2) Consistency regularization methods. Consistency regularization methods impose consistency restrictions on the final loss functions in accordance with the smoothness or manifold assumptions. Three viewpoints, namely input perturbations, weights perturbations, and layer perturbations of the network, may be used to build constraints. The Teacher-Student model~\cite{tarvainen2017mean} is the consistency regularization method architecture that has been used most frequently~\cite{sajjadi2016regularization,ke2019dual}.

(3) Graph-based methods. The fundamental premise of graph-based learning is the construction of a similarity graph from the raw dataset, where each node represents a training sample and each weighted edge indicates the degree of similarity between two nodes. Based on the several assumptions, the built graph may be used to infer the label information of unlabeled samples~\cite{gilmer2017neural,zhou2020towards}.

(4) Self-training methods. The most prevalent approach of self-training methods is to produce pseudo labels for unlabeled instances in accordance with the high confidence model's prediction and then use them to regularize the model's training~\cite{rosenberg2005semi,dong2018tri}. These methods could also be treated as bootstrapping algorithms.

(5) Hybrid methods. To increase performance, hybrid methods include a variety of procedures, including consistency regularization, pseudo-labeling, data augmentation, entropy estimation, and other elements. Among the methods represented are Mixup~\cite{zhang2018mixup}, MixMatch~\cite{berthelot2019mixmatch}, ReMixMatch~\cite{berthelot2020remixmatch}, FlxMatch~\cite{sohn2020fixmatch}, FlexMatch~\cite{zhang2021flexmatch}, etc.

Our {\gradient} ({\modelname}) method belongs to self-training methods which could incrementally assign pseudo labels to unlabeled data and leverage these pseudo labels to iteratively improve the classification capability of the model. However, these methods always endure gradual drift problem \cite{zhang2016understanding,liu2021noisy}: during the training process, the generated pseudo label data contains noise and could not been corrected through the model itself. Using these pseudo label data iteratively cause the model to deviate from the global minima. Our work alleviates this problem by encouraging pseudo-labeled data to imitate the gradient optimization direction on the labeled data, and introducing an effective feedback loop to improve generalization ability via reinforcement learning.

\subsection{Reinforcement Learning in Nature Language Processing}
Reinforcement Learning is widely used in Nature Language Processing \cite{narasimhan2016improving,li2016deep,takanobu2019hierarchical}. These methods are all designed with rewards to force the correct actions to be executed during the model training process, so as to improve model performance. In this paper, we divide reinforcement learning in NLP tasks into four main groups \cite{uc2022survey}.

(1) Syntactic parsing. Analyzing a string of symbols from some alphabet is known as syntactic parsing. Such analysis is frequently carried out in accordance with a set of rules known as grammar. Many parsing trees may be produced by a grammar, and each of these trees provides the proper construction for sentences in the associated language. Reinforcement learning techniques are especially well adapted for the underlying sequential choice issue since parsing can be seen as a sequential search problem with a parse tree as the ultimate target state. When an optimum policy is employed in a particular Markov decision process, a parse is often produced as a route \cite{neu2009training,le2017tackling,cao2019semantic}.

(2) Language understanding. Since the language understanding task can also be modeled as a Markov decision process, we could adopt sophisticated reinforcement learning algorithms designed recently \cite{he2016deep}. Additionally, they may be used in conjunction with deep neural networks to handle the large amounts of data that text understanding applications often need \cite{guo2017learning,zhu2020dual}.

(3) Text generation. The goal of text generation is to mechanically produce valid natural language sentences. A language model is one of these systems' components. The optimization challenge is to come up with valid sequences of substrings through reinforcement learning algorithms that will eventually finish a whole sentence when the language model is given or learnt \cite{keneshloo2019deep,li2018paraphrase}.

(4) Machine translation. Large neural networks are used by neural machine translation (NMT) to forecast the likelihood of a word sequence \cite{stahlberg2020neural}. Modern phrase-based NMT systems, where a unit of translation may be a series of words, have benefited from the widespread application of NMT approaches \cite{hassan2018achieving,lam2019interactive}. During training of seq2seq models, two problems arise: exposure bias and inconsistency between the training and test objectives. Both problems have recently been studied, and several reinforcement learning solutions have been proposed \cite{keneshloo2019deep}.


Our {\gradient} ({\modelname}) method belongs to the category of language understanding tasks. In our work, we define reward as the cosine similarity between gradient vectors calculated from pseudo-labeled data and labeled data. 


\section{Vanilla Models}\label{sec:basemodel}
In this section, we first introduce the three sub-tasks: Named Entity Recognition, Relation Extraction, and Event Extraction in detail. As illustrated in Figures \ref{fig:vanilla_ner}, \ref{fig:vanilla_re}, and \ref{fig:vanilla_ee}, since the forms of the three tasks are different, we design the vanilla models respectively, and based on the vanilla model for low-resource scenarios, we leverage {\modelname} to prove its effectiveness for general low-resource information extraction tasks.

\subsection{Named Entity Recognition}\label{sec:ner}
Named Entity Recognition (NER) aims to infer a label for each token to indicate whether it belongs to an entity and classify entities into predefined types. Typically a named entity refers to a word or phrase that serves as a proper name for something or someone \cite{petasis2000automatic}. Common named entity types include Person, Organization, and Location, etc.

The named entity recognition task could be formulated as follows: Given an input sentence, the goal of named entity recognition is to predict named entity triples $\left\langle s, e, t\right\rangle$ in the sentence, where $s$ is the start token index, $e$ is the end token index and $t$ is the associated entity type (from a predefined type set) of the named entities.

\begin{figure}[bt!]
    \centering
    \includegraphics[width=1\linewidth]{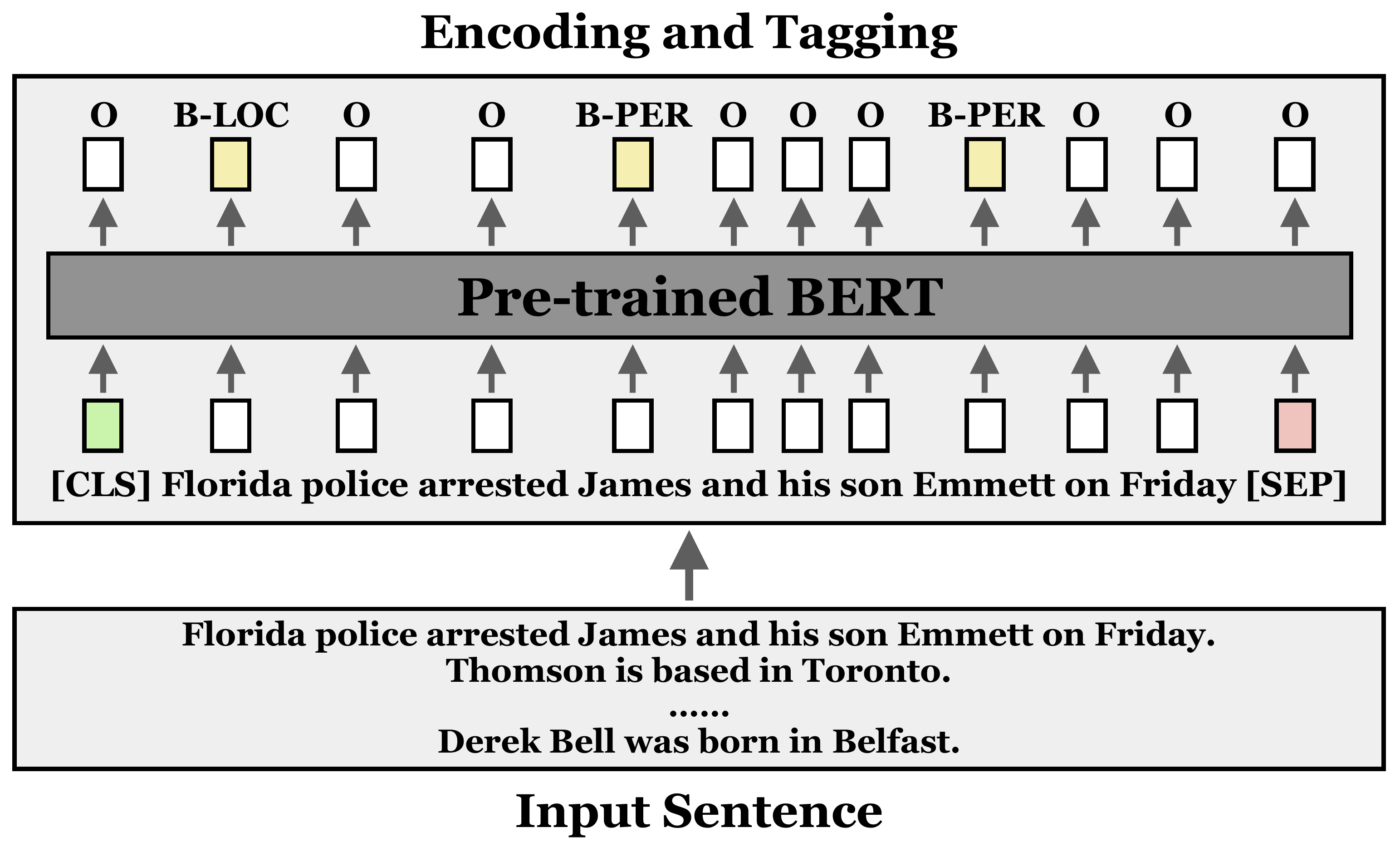}
    \caption{Overview of vanilla model for named entity recognition tasks.}
    \label{fig:vanilla_ner}
\end{figure}

As shown in Figure \ref{fig:vanilla_ner}, the vanilla model recognizes named entities in the input sentence in two steps: encoding and tagging. Now, we give the details of these two steps.

\subsubsection{Encoding}

Given an input sentence ${\left[x_{1}, x_{2}, \ldots, x_{L}\right]}$ of $L$ tokens after the tokenizer, we feed the sentence into a pre-trained language model (BERT) to obtain contextualized embeddings $\boldsymbol{h}_{i}$ for each token ${x_{i}}$:
\begin{equation}
\boldsymbol{H}=\left[\boldsymbol{h}_{1}, \boldsymbol{h}_{2}, \ldots, \boldsymbol{h}_{L}\right]=\operatorname{BERT}\left(\left[x_{1}, x_{2}, \ldots, x_{L}\right]\right).
\end{equation}

\subsubsection{Tagging}

The task of named entity recognition could be seen as a sequence tagging task in which we assign a label $y_{i}$ to each token $x_{i}$ in the input sentence with \texttt{BIO} tags. In \texttt{BIO} tagging, we label any token that begins a span of an entity with the prefix \texttt{B-} (eg, \texttt{B-PER}) and that occurs inside a span with the prefix \texttt{I-} (eg, \texttt{I-PER}). Any token outside of the span of entity will be labeled as \texttt{O}. 
We adopt two hidden MLP layers for tagging:
\begin{equation}
\hat{\boldsymbol{y}}_{i}=\boldsymbol{W}_{o}\left(\boldsymbol{W}_{h}\boldsymbol{h}_{i}+\boldsymbol{b}_{h}\right)+\boldsymbol{b}_{o},
\end{equation}
where $\boldsymbol{W}_{h}$, $\boldsymbol{W}_{o}$, $\boldsymbol{b}_{h}$ and $\boldsymbol{b}_{o}$ are learnable parameters, and $\hat{\boldsymbol{y}}_{i}$ is the tag logits of the token $x_{i}$. The training objective is to minimize the following loss function:
\begin{equation}
\mathcal{L}_{NER}=-\frac{1}{N} \sum_{i=1}^{N} \sum_{j=1}^{L_{i}} \boldsymbol{y}_{j}^{i} \log \hat{\boldsymbol{y}}_{j}^{i},
\end{equation}
where $N$ is the size of training set, $L_{i}$ is the length of the $i$-th sentence in the training set, and $\boldsymbol{y}_{j}^{i}$ is the ground-truth tag vector of the $j$-th token in the $i$-th sentence.
\begin{figure}[bt!]
    \centering
    \includegraphics[width=0.87\linewidth]{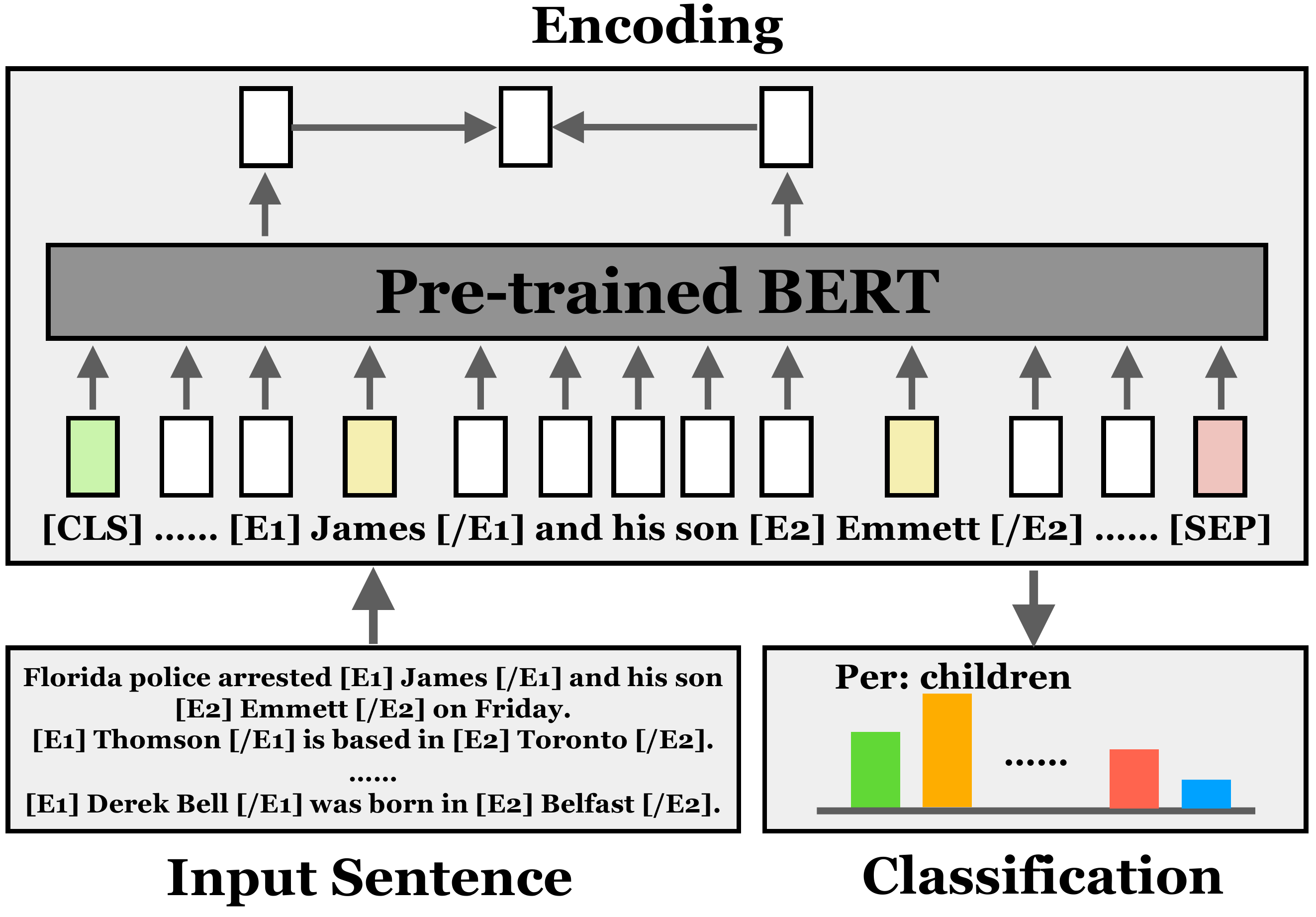}
    \caption{Overview of vanilla model for relation extraction tasks.}
    \label{fig:vanilla_re}
\end{figure}

\subsection{Relation Extraction}\label{sec:ee}
Relation Extraction attempts to extract triplets of the form (head entity, relation, tail entity) from sentences. The extracted triplets from the sentence could be adopted in various downstream applications like recommendation systems, question answering, and natural language understanding.

The relation extraction task could be formulated as follows: Given an input sentence, the goal of relation extraction is to predict relational triples $\left\langle entity_{head}, r, entity_{tail}\right\rangle$ in the sentence, where head and tail entities have been recognized in advance via named entity recognition task.

As depicted in Figure \ref{fig:vanilla_re}, the vanilla model extracts triplets from the input sentence in two steps: encoding and classification. Now, we give the details of these two steps.

\subsubsection{Encoding}

Given an input sentence ${\left[x_{1}, x_{2}, \ldots, x_{L}\right]}$ of $L$ tokens after the tokenizer, we introduce four special tokens ${\left[E_{1}\right]}$, ${\left[/E_{1}\right]}$, ${\left[E_{2}\right]}$, ${\left[/E_{2}\right]}$ and insert them into the sentence to mark the beginning and the end of entities \cite{soares2019matching}. Then we feed the sentence into a pre-trained language model (BERT) to obtain the contextualized embeddings $\boldsymbol{h}_{i}$ for each token ${x_{i}}$:

\begin{equation}
\begin{aligned}
\boldsymbol{H}=\left[\boldsymbol{h}_{1}, \boldsymbol{h}_{2}, \ldots, \boldsymbol{h}_{L+4}\right]=\operatorname{BERT}\big([x_{1},...,[E_{1}],x_{i},\\...,x_{j-1},[/E_{1}],...,[E_{2}],x_{k},...,x_{m-1},[/E_{2}],...,x_{L}]\big).
\end{aligned}
\end{equation}

To obtain the relational embedding of two entities ${\left[E_{1}\right]}$ and ${\left[E_{2}\right]}$, instead of using sentence-level semantics, as illustrated in Figure \ref{fig:vanilla_re}, we adopt the output embeddings corresponding to ${\left[E_{1}\right]}$ , ${\left[E_{2}\right]}$ positions and concatenate them to derive a fixed-length relational embedding:
$
\boldsymbol{h}^{r} = [\boldsymbol{h}_{[E_{1}]}, \boldsymbol{h}_{[E_{2}]}].
$
\subsubsection{Classification}
Similar to the tagging process of the NER task, we adopt two hidden layers for contextualized relational embedding classification:
\begin{equation}
\hat{\boldsymbol{y}}_{i}=\boldsymbol{W}_{o}\left(\boldsymbol{W}_{h}\boldsymbol{h}_{i}+\boldsymbol{b}_{h}\right)+\boldsymbol{b}_{o},
\end{equation}
where $\boldsymbol{W}_{h}$, $\boldsymbol{W}_{o}$, $\boldsymbol{b}_{h}$ and $\boldsymbol{b}_{o}$ are learnable parameters, and $\hat{\boldsymbol{y}}_{i}$ is the relational label logits of the $i$-th sentence. The training objective is to minimize the following loss function:
\begin{equation}
\mathcal{L}_{RE}=-\frac{1}{N} \sum_{i=1}^{N} \boldsymbol{y}^{i} \log \hat{\boldsymbol{y}}^{i},
\end{equation}
where $N$ is the size of the training set, $\boldsymbol{y}^{i}$ is the ground-truth relational label vector of the $i$-th sentence.

\begin{figure}[bt!]
    \centering
    \includegraphics[width=1\linewidth]{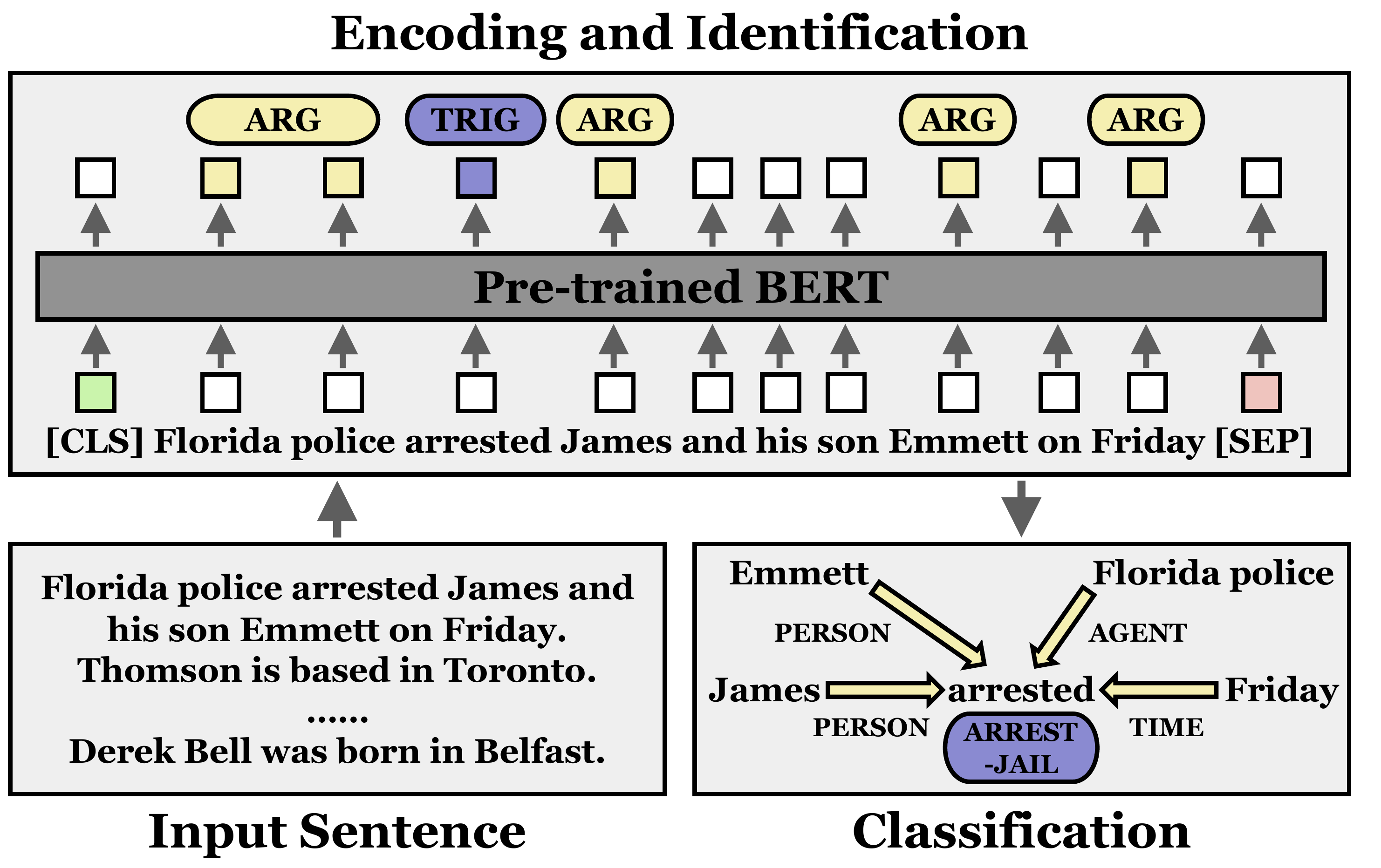}
    \caption{Overview of vanilla model for event extraction tasks.}
    \label{fig:vanilla_ee}
\end{figure}
\subsection{Event Extraction}\label{sec:ee}

Event extraction aims to identify event triggers and their arguments in unstructured texts and classify them into pre-defined types (for event triggers) and roles (for event arguments). Event triggers are the main words or phrases that most clearly express an event occurrence, typically a verb or a noun. Event arguments refer to the words or phrases that serve as a participant or attribute with a specific role in those events \cite{ace2005guideline,lu2021text2event}.

The event extraction task could be formulated as follows \cite{lin2020joint}: Given an input sentence, the goal of event extraction is to predict an event graph $G=(V, E)$, where $V$ and $E$ are the node and edge sets, respectively. Each node $v_{i}=\left\langle a_{i}, b_{i}, l_{i}\right\rangle \in V$ represents an entity mention or event trigger, where $a$ is the start token index, $b$ is the end token index and $l$ is the associated node type label (entity type for entity mentions and event type for event triggers). Each edge $e_{i j}=\left\langle i, j, l_{i j}\right\rangle \in E$ represents the event-argument link between $v_{i}$ and $v_{j}$, where $l_{i j}$ is the argument role of $v_{j}$ in $v_{i}$.

As illustrated in Figure \ref{fig:vanilla_ee}, the base model used for event extraction extracts the event graph from a given sentence in three steps: encoding, identification, and classification. Now, we give the details of these three steps.

\subsubsection{Encoding}
Given an input sentence of $L$ tokens after the tokenizer, we feed the sentence into a pre-trained language model (BERT) to obtain the contextualized embeddings $\boldsymbol{h}_{i}$ for each token:
\begin{equation}
\boldsymbol{H}=\left[\boldsymbol{h}_{1}, \boldsymbol{h}_{2}, \ldots, \boldsymbol{h}_{L}\right]=\operatorname{BERT}\left(\left[x_{1}, x_{2}, \ldots, x_{L}\right]\right).
\end{equation}

\subsubsection{Identification}
To identify the entity mentions and event triggers in the sentence which act as nodes in the event graph, we use two separate taggers for identification. We assign each token a tag with \texttt{BIO} scheme. We map each token to its corresponding score vector with a linear layer. Then we use a conditional random field layer to capture the dependencies between two adjacent predicted tags in the tag path. The whole process is formulated as:
\begin{equation}
\hat{\boldsymbol{y}}_{i}=\boldsymbol{W}_{I}\boldsymbol{h}_{i}+\boldsymbol{b}_{I},
\end{equation}
\begin{equation}
s(\boldsymbol{H}, \hat{\boldsymbol{z}})=\sum_{i=1}^{L} \hat{y}_{i, \hat{z}_{i}}+\sum_{i=1}^{L+1} A_{\hat{z}_{i-1}, \hat{z}_{i}},
\end{equation}
where $\boldsymbol{W}_{I}$, $\boldsymbol{b}_{I}$ and $\boldsymbol{A}$ are model parameters whose values are learned during training, $\hat{\boldsymbol{y}}_{i}$ is the tag score vector of the token $x_{i}$, $\hat{\boldsymbol{z}}=\left\{\hat{z}_{1}, \ldots, \hat{z}_{L}\right\}$ is a tag path of the input sequence, $\hat{y}_{i, \hat{z}_{i}}$ is the $\hat{z}_{i}$-th component of the score vector $\hat{\boldsymbol{y}}_{i}$ and $A_{\hat{z}_{i-1}, \hat{z}_{i}}$ is the $\left(\hat{z}_{i-1}, \hat{z}_{i}\right)$ entry in the transition matrix $\boldsymbol{A}$ which represents the transition score from tag $\hat{z}_{i-1}$ to $\hat{z}_{i}$. We add two spacial tags $<$b$>$ and $<$e$>$ to the tag path as $\hat{z}_{0}$ and $\hat{z}_{L+1}$ to denote the beginning and end of the sequence.

During training phase, we maximize the log-likelihood of the ground-truth tag path as:
\begin{equation}
\log p(\boldsymbol{z} \mid \boldsymbol{H})=s(\boldsymbol{H}, \boldsymbol{z})-\log \sum_{\hat{\boldsymbol{z}} \in Z} e^{s(\boldsymbol{H}, \hat{\boldsymbol{z}})},
\end{equation}
where $Z$ is the set of all possible tag paths for a given sequence. Based on the log-likelihood, we obtain the first part of the total loss, which involves identification as
\begin{equation}
\mathcal{L}^{\mathrm{I}}=-\log p(\boldsymbol{z} \mid \boldsymbol{H}).
\end{equation}

\subsubsection{Classification}
We do not use the predicted types but only the spans of entity mentions or event triggers at the identification stage. Instead, we classify the type of entity mentions and event triggers, as well as the role of arguments (an entity mention in a specific event), at classification stage. Each node identified in the previous step is represented as $\boldsymbol{v}_{i}$ by averaging its word embeddings. Then we use separate task-specific linear layers to calculate label scores for each node or edge as:
\begin{equation}
\hat{\boldsymbol{y}}_{i}^{c}=\boldsymbol{W}_{o}^{c}\left(\boldsymbol{W}_{h}^{c}\boldsymbol{v}_{i}+\boldsymbol{b}_{h}^{c}\right)+\boldsymbol{b}_{o}^{c},
\end{equation}
\begin{equation}
\hat{\boldsymbol{y}}_{k}^{c}=\boldsymbol{W}_{o}^{c}\left(\boldsymbol{W}_{h}^{c}\left[\boldsymbol{v}_{i}, \boldsymbol{v}_{j}\right]+\boldsymbol{b}_{h}^{c}\right)+\boldsymbol{b}_{o}^{c},
\end{equation}
respectively, where $\boldsymbol{W}_{h}^{c}$, $\boldsymbol{W}_{o}^{c}$, $\boldsymbol{b}_{h}^{c}$ and $\boldsymbol{b}_{o}^{c}$ are model parameters, and $c$ indicates a specific task among entity mention type classification, event trigger type classification, and argument role classification.

For each classification task, the training objective is to minimize the following cross-entropy loss, which is the second part of the total loss involving classification:
\begin{equation}
\mathcal{L}^{\mathrm{c}}=-\frac{1}{N^{c}} \sum_{i=1}^{N^{c}} \boldsymbol{y}_{i}^{c} \log \hat{\boldsymbol{y}}_{i}^{c}
\end{equation}
where $\boldsymbol{y}_{i}^{c}$ is the ground-truth label vector and $N^{c}$ is the number of instances for the task $c$.

During training, we optimize the following total objective function:
\begin{equation}
\mathcal{L}_{EE}=\mathcal{L}^{I}+\sum_{c \in C} \mathcal{L}^{c}.
\end{equation}
where $C$ is the set of classification tasks. During the inference stage, as shown in Figure \ref{fig:vanilla_ee}, we simply predict the label with the highest score for each node and edge to generate the locally best graph ${G}$.
\section{Proposed Framework}\label{sec:model}
In this section, we first introduce the proposed framework Gradient Imitation Reinforcement Learning ({\modelname}), which is a simple, generic yet powerful technique. For general IE sub-tasks (eg, NER, RE, and EE), {\modelname} can flexibly adapt to the limited labeled data of each sub-task, and improve the model generalization ability beyond existing human annotations.

As illustrated in Figure \ref{fig:overview}, the inputs to the vanilla models for the three IE sub-tasks are limited labeled data and large amounts of unlabeled data. 
In a traditional self-training setting, we fine-tune vanilla models directly on the labeled data, and let vanilla models assign pseudo labels on unlabeled data as pseudo-labeled data. However, we argue that such learning paradigm suffers from selection bias due to the lack of feedback loops: the bias occurs when a model itself influences the generation of data which are later used for training.
In this work, we complete the feedback loop and alleviate such bias by leveraging {\modelname} to learn a policy that maximizes the likelihood between the expected gradient optimization direction from pseudo-labeled data, and the standard gradient optimization direction on labeled data.
\definecolor{applegreen}{rgb}{0.55, 0.71, 0.0}
\definecolor{babyblueeyes}{rgb}{0.63, 0.79, 0.98}

\begin{figure}[bt!]
    \centering
    \includegraphics[width=\linewidth]{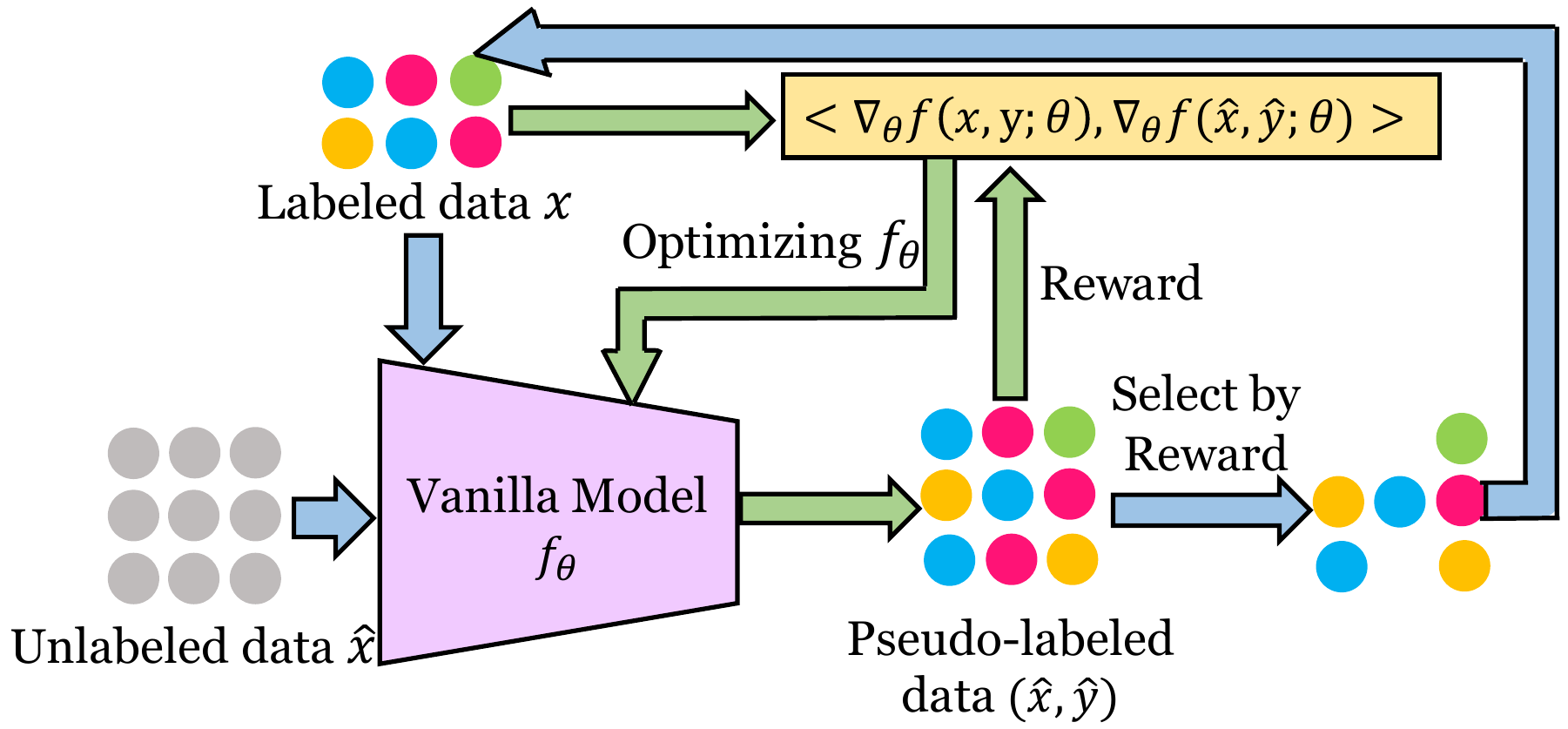}
    \caption{Overview of the proposed {\modelname} framework for Low-Resource Information Extraction. {\color{babyblueeyes}{Blue}} arrows represent the flow of data and {\color{applegreen}{green}} arrows represent Gradient Imitation Reinforcement Learning. }
    \label{fig:overview}
\end{figure}
Specifically, we assign pseudo labels via vanilla model on unlabeled data as pseudo-labeled data, and add the selected pseudo-labeled data into the existing labeled data to iteratively improve vanilla model. 
We argue that without a feedback loop measuring the quality of pseudo labels, the model is more likely to suffer from selection bias and is impeded towards a better generalization ability.

We aim to generate pseudo labels with fewer labeling biases and errors, especially with scarce annotations. 
To achieve this goal, we focus on improving the vanilla model performance by introducing gradient imitation to define and quantify what an appealing behavior looks like. We define the partial derivatives of the loss function corresponding to the vanilla model parameters on the labeled data as standard gradient descent. We assume that when pseudo-labeled data are correctly labeled in vanilla model, partial derivatives to the vanilla model parameters on the pseudo-labeled data would be highly similar to standard gradient descent. So we maximize the correlation between gradients over the pseudo-labeled data and those over the labeled data. Following this assumption, we propose Gradient Imitation Reinforcement Learning ({\modelname}), which optimizes vanilla model under a reinforcement learning framework \cite{williams1992simple}. Now, we explain the reinforcement learning process in detail.

\noindent \textbf{State:}

State is used to signal the optimization status. We use ${\boldsymbol{s}^{(t)}}$ to denote the state. ${\boldsymbol{s}^{(t)}}$ consists of the updated labeled dataset ${\mathcal{D}_{l}}$ at step $t$, along with a standard gradient direction $\boldsymbol{g}_{l}$ at step $t$. We will give the definition of $\boldsymbol{g}_{l}$ in Reward part.

\noindent\textbf{Policy:}

Our policy is learned to assign correct pseudo labels on unlabeled data. As described in Section \ref{sec:basemodel}, for the three IE sub-tasks, we adopt three vanilla models to pseudo-label the unlabeled data of the corresponding sub-tasks. Therefore, for each sub-task, the policy network is parameterized separately by the corresponding vanilla model network $f_{\theta}$. Note that we use $f_{\theta}$ to replace the three vanilla models below for a unified representation.

\noindent\textbf{Action:}

The action is to predict label on unlabeled data ${\boldsymbol{\hat{x}}^{(t)}}$ as pseudo-labeled data ${(\boldsymbol{\hat{x}}^{(t)},\boldsymbol{\hat{y}}^{(t)})}$ given the State at step $t$. We consider the predicted label that corresponds to the Policy network of three IE sub-tasks as the pseudo label ${\boldsymbol{\hat{y}}^{(t)}}$:
\begin{equation}
\boldsymbol{\hat{y}}^{(t)}=f_{\theta}(\boldsymbol{\hat{x}}^{(t)}).
\end{equation}

\noindent\textbf{Reward:}

We use reward to signal labeling biases from the current policy on pseudo-labeled data. Our goal is to minimize the approximation error of the gradients obtained over the pseudo-labeled data. In other words, we maximize the correlation between gradients over the pseudo-labeled data and those over the labeled data. 

We define the standard gradient descent direction on the all ${N}$ labeled data as ${\boldsymbol{g}_{l}}$ and the expected gradient descent direction on the pseudo-labeled data as ${\boldsymbol{g}_{p}}$ respectively:
\begin{align}
&{\boldsymbol{g}_{l}}^{(n)}(\theta)=\nabla_{\theta} \mathcal{L}_{l}\left(\boldsymbol{x}^{(n)},\boldsymbol{y}^{(n)} ; \theta\right),\\
&{\boldsymbol{g}_{p}}^{(t)}(\theta)={\nabla_{\theta}} \mathcal{L}_{p}\left( \boldsymbol{\hat{x}}^{(t)},\boldsymbol{\hat{y}}^{(t)}; \theta\right),
\end{align}
where 
${\nabla_{\theta}}$ refers to the partial derivatives of the optimization loss $\mathcal{L}$ corresponding to Policy ${f_{\theta}}$ with respect to ${{\theta}}$. 
Considering that the outliers in the labeled data will affect the direction of standard gradient descent, we approximate ${g_{l}}$ over all ${N}$ labeled data and we define $\mathcal{L}_{l}$ (labeled data) and $\mathcal{L}_{p}$ (pseudo-labeled data) as:
\begin{align}
\mathcal{L}_{l} =& \frac{1}{N}\sum_{n=1}^{N}\mathit{loss}(f_{\theta}({\boldsymbol{x}}^{(n)}),\boldsymbol{y}^{(n)}),\\
\label{generated_loss}\mathcal{L}_{p} &= \mathit{loss}(f_{\theta}(\boldsymbol{\hat{x}}^{(t)}),\boldsymbol{\hat{y}}^{(t)}),
\end{align}
where $\mathit{loss}$ is $\mathcal{L}_{NER}$, $\mathcal{L}_{RE}$, and $\mathcal{L}_{EE}$ respectively and ${\boldsymbol{y}^{(n)}}$ is a one-hot vector indicating the target label assignment.

Since the most important guidance obtained by the gradient vector ${\boldsymbol{g}_{l}}$ is its gradient descending direction, we measure the discrepancy between ${\boldsymbol{g}_{l}}$ and ${\boldsymbol{g}_{p}}$ for state ${\boldsymbol{s}^{(t)}}$ by defining their cosine similarity 
as the reward:
\begin{align}
R^{(t)}=\frac{\boldsymbol{g}_{l}(\theta)^{\mathrm{T}} \boldsymbol{g}_{p}(\theta)}{\left\|\boldsymbol{g}_{l}(\theta)\right\|_{2}\left\|\boldsymbol{g}_{p}(\theta)\right\|_{2}}.
\end{align}
The range of ${R^{(t)}}$ is [-1,1]. For those pseudo-labeled data ${(\boldsymbol{\hat{x}}^{(t)},\boldsymbol{\hat{y}}^{(t)})\in\mathcal{D}_{p}}$ with ${R^{(t)}>\lambda}$, ${\lambda=0.5}$, we treat them as positive reinforcement to improve the generalization ability of vanilla model network. We add these selected pseudo-labeled data to the labeled data and correct the standard gradient descending direction:
\begin{align}
&\mathcal{D}_{l}\leftarrow \mathcal{D}_{l}\cup\mathcal{D}_{p},\\
\label{eq:standard gradient} \boldsymbol{g}_{l}\leftarrow& \frac{1}{N+1}(N\boldsymbol{g}_{l}+\boldsymbol{g}_{p}).
\end{align}
For Eq. \eqref{eq:standard gradient}, we set the weight of the updated gradient direction according to the number of samples, where the standard gradient direction is calculated using all $N$ labeled samples and each pseudo labeled sample. The positive feedback obtained from {\modelname} via trial and error can attribute the improvement of vanilla model network (Policy) to assign correct pseudo label for next unlabeled data ${\boldsymbol{\hat{x}}^{(t)}}$ (State). 

\noindent\textbf{Reinforcement Learning Loss:}

We adopt the REINFORCE algorithm \cite{williams1992simple} and Policy Gradient for optimization. We calculate the loss over a batch of pseudo-labeled samples. The vanilla model will be optimized by {\modelname} on each batch according to the following reinforcement learning loss:
\begin{align}
\begin{split}
\mathcal{L}(\theta) = \sum_{t=1}^{T} \mathit{loss} \big(f_\theta(\boldsymbol{\hat{x}}^{(t)}),\boldsymbol{\hat{y}}^{(t)}\big)*R^{(t)},
\end{split}
\end{align}
where $\mathit{loss}$ can be $\mathcal{L}_{NER}$, $\mathcal{L}_{RE}$, or $\mathcal{L}_{EE}$ depending on type of sub-tasks. ${R^{(t)}}$ is the reward and ${\boldsymbol{\hat{y}}^{(t)}\sim\pi(\cdot|\boldsymbol{\hat{x}}^{(t)};\theta)}$. The $\pi$ function means Policy in reinforcement learning. In our setting, it is parameterized as $f_{\theta}$, which is learned to assign pseudo labels on unlabeled data and we minimize ${\mathcal{L}(\theta)}$ to optimize the ${\theta}$. $T$ represents a total number of time steps in a reinforcement learning episode and is set to $16$, the same number as the batch size. For each pseudo-labeled data with high reward (${R^{(t)}>\lambda}$, ${\lambda=0.5}$), we use it to dynamically update the labeled dataset \& standard gradient direction and guide the reinforcement learning process to the next State.

Note that ${f_{\theta}}$ is first pretrained using limited labeled data in a supervised way. During the process of calculating reinforcement learning loss, our model follows the \textbf{Markov decision process} and the labeled data ${\mathcal{D}_{l}}$ and standard gradient descending direction ${\boldsymbol{g}_{l}}$ will be dynamically corrected by the selected pseudo-labeled data ${\mathcal{D}_{p}}$, which means that for each State, Policy will be updated over time $t$. The vanilla model could solicit positive feedbacks obtained using {\modelname} via trial and error.

\noindent\textbf{Adapt {\modelname} to Vanilla Models:}

In the implementation process, we could directly adapt the general framework {\modelname} to the three vanilla models $f_{\theta}$ without any modification, which also reflects the flexibility and generalization of {\modelname} for general IE tasks.

\section{Experiments}\label{sec:experiments}

We conduct extensive experiments on seven public datasets across three IE sub-tasks (named entity recognition, relation extraction, and event extraction) to demonstrate the effectiveness of our Gradient Imitation Reinforcement Learning for low-resource IE in both semi-supervised IE and few-shot IE settings. We also provide a detailed analysis of each module to show the advantages of {\modelname}.

\subsection{Datasets}
\subsubsection{Named Entity Recognition}
We evaluate our framework on three widely-adopted public datasets: \textbf{OntoNotes 5.0} \cite{pradhan2013towards}, \textbf{CoNLL-2003} \cite{sang2003introduction}, and \textbf{WNUT 2017} \cite{derczynski2017results}. OntoNotes 5.0 is a large-scale, multi-genre corpus manually annotated with syntactic and semantic information including named entity annotations. There are totally 75181, 9603, and 9479 training, validation, test sentences and 18 named entity types such as PERSON, PRODUCT and DATE. CoNLL-2003 is a named entity recognition dataset released as a part of CoNLL-2003 shared task: Language-independent named entity recognition, which concentrates on 4 types of named entities including Persons, Locations, Organizations, and Miscellaneous. The size of training, validation, and test set is 14041, 3250, and 3453 respectively. WNUT 2017 is the named entity recognition dataset of the WNUT 2017 Emerging and Rare entity recognition shared task, which focuses on identifying unusual, previously-unseen entities in the context of emerging discussions. WNUT 2017 has 3394, 1009, and 1287 training, validation and test instances respectively, with 6 named entity types: Person, Location, Corporation, Consumer good, Creative work, and Group.


\subsubsection{Relation Extraction}
Following previous work \cite{hu2020semi,li2020exploit}, we conduct experiments on two public RE datasets, including the SemEval 2010 Task 8 (\textbf{SemEval}) \cite{hendrickx2010semeval}, and the TAC Relation Extraction Dataset (\textbf{TACRED}) \cite{zhang2017position}. SemEval is a standard benchmark dataset for evaluating relation extraction models, which consists of training, validation, and test set with 7199, 800, and 1864 relation mentions, respectively, and also with 19 relations types in total (including \textit{no\_relation}), of which the \textit{no\_relation} percentage is 17.4\%. TACRED is a large-scale crowd-sourced relation extraction dataset that is collected from all previous TAC KBP relation schemas. The dataset consists of training, validation, and test set with 75049, 25763, and 18659 relation mentions respectively, with 42 relation types in total (including \textit{no\_relation}), of which \textit{no\_relation} percentage is 78.7\%.

We observe that TACRED is far more complicated than SemEval, since it has more relation types and more skewed distribution among different relation categories. Note that the name entities in the sentences have been recognized and marked in advance.

\subsubsection{Event Extraction}
Following the previous work \cite{lin2020joint}, we perform our experiments on two benchmark event extraction datasets, including the \textbf{ACE05-E} and \textbf{ACE05-E$^+$}. These two datasets are derived from the Automatic Content Extraction 2005 Multilingual Training Corpus \cite{ace2005guideline}, which contains approximately 1800 files of mixed genre texts annotated for entities, relations, and events. The genres include newswire, broadcast news, broadcast conversation, weblog, discussion forums and conversational speech. Following the preprocessing in\cite{wadden2019entity}, ACE05-E includes named entity, relation and event annotations, where it keeps 7 entity types, 33 event types and 22 argument roles. ACE05-E$^+$ is a revised event extraction dataset based on ACE05-E, proposed by Lin \textit{et al.} \cite{lin2020joint}. ACE05-E$^+$ adds back pronouns, multi-token event triggers, and the order of relation arguments to reinstate some important elements that are absent from ACE05-E. Furthermore, the lines before the $<$text$>$ tag like headline and datetime are skipped due to the lack of annotation.

\subsection{Baselines and Evaluation metrics}
Our three vanilla models are flexible to integrate different contextualized encoders during encoding step. From Tables \ref{tab:main_ner}, \ref{tab:main_re}, \ref{tab:main_ee}, and \ref{tab:main_ee_2}, we first compare several widely used supervised encoders with only labeled data: \textbf{LSTM} \cite{hochreiter1997long}, \textbf{PCNN} \cite{zeng2015distant}, \textbf{PRNN} \cite{zhang2017position}, and \textbf{BERT} \cite{devlin2019bert}. Among them, BERT achieved state-of-the-art performance. So we adopt pre-trained BERT as the base encoder for three vanilla models and other baselines for a fair comparison.

For baselines, as illustrated in Section \ref{sec:related}, existing deep low-resource information extraction methods could be divided into five categories.
We compare {\modelname} with seven methods that are representative or SOTA in each category for comparison:

\begin{table*}[t]
\centering
\caption{F1 (\%) comparisons on the OntoNotes 5.0, CoNLL-2003, and WNUT 2017 datasets with various amounts of labeled data and 50\% unlabeled data for NER task.}
\label{tab:main_ner}
  \resizebox{\textwidth}{!}{
\begin{tabular}{lcccccccccccc}
\thickhline
\multicolumn{1}{l}{\multirow{2}{*}{\textbf{Methods / \%Labeled Data}}} & \multicolumn{4}{c}{\textbf{OntoNotes 5.0}}& \multicolumn{4}{c}{\textbf{CoNLL-2003}} & \multicolumn{4}{c}{\textbf{WNUT 2017}} \\ \cmidrule(lr){2-5}\cmidrule(lr){6-9}\cmidrule(lr){10-13}
\multicolumn{1}{c}{}     & 5-shot     & 10-shot   & 5\%    & 10\%  & 5-shot     & 10-shot   & 5\%    & 10\%    & 5-shot   & 10-shot  & 3\%    & 10\%     \\ \hline
                           LSTM \cite{hochreiter1997long}     & 34.29±3.79   &   39.82±3.51     &     47.29±3.26     &  50.07±3.08       &  40.58±4.02   &  47.08±2.98    &  60.44±2.45   &  62.08±2.07    &    8.24±3.17     &  12.67±2.50  &12.63±2.24 & 21.36±2.23  \\
                          PCNN \cite{zeng2015distant}       &  36.84±3.78  &  42.76±3.04   &   54.38±2.74   &  56.78±3.42  &  43.82±3.94   & 49.22±2.16  & 69.55±2.62 & 71.26±1.87      &  9.36±3.57    &  14.80±2.71 &14.92±2.72 &20.87±1.94  \\
                           PRNN \cite{zhang2017position}     &  42.17±2.66  &  51.09±2.17   &  62.59±3.08 & 66.72±2.80  & 50.12±2.72 & 56.01±1.94   & 73.42±1.38 & 74.50±1.33  &  12.59±3.82  &  21.44±2.56 & 21.42±2.04 & 27.26±1.88 \\
                          BERT \cite{devlin2019bert}   &  53.43±0.68  &  64.83±1.00    &  77.42±0.79  &  80.73±1.04   & 56.55±0.93 &  65.08±1.04  & 82.51±1.00 & 83.60±0.67  & 20.17±1.07 &  28.13±0.98 &
                          28.20±1.24 &
                          36.09±1.71  \\\hline \hline
                       BiGAN$_{BERT}$ \cite{donahue2016adversarial}   &  54.92±0.97  & 65.74±1.80  & 78.81±0.83 & 81.94±0.98 & 58.82±1.40  & 66.83±1.53  & 83.04±1.28 &  84.57±1.70   &  21.21±1.02  &  30.72±1.04 & 30.76±1.21 &36.09±1.71 \\
                        Mean Teacher$_{BERT}$ \cite{tarvainen2017mean}  &  56.33±0.88  &  66.13±0.68   & 79.65±0.54   & 83.14±1.07 & 60.24±1.28  & 67.84±1.42   & 84.19±1.03 &  85.92±1.36  &  22.07±1.13  & 31.52±1.00 &31.58±0.98 &40.20±1.51\\
                     UDA$_{BERT}$ \cite{xie2020unsupervised}  &  56.42±0.79  &  66.45±0.89  &  79.83±0.58   & 83.32±1.21 & 60.58±1.62 & 68.03±1.21  & 84.27±0.84 & 86.24±1.29   &  22.51±0.82    &  31.74±0.83 &31.79±1.02 & 40.37±1.44\\
                      Deep Co-training$_{BERT}$ \cite{blum1998combining}  & 56.44±1.24  &  66.23±0.90  & 79.87±0.60 &  83.16±0.80   & 60.01±1.00 & 68.14±1.36  & 84.03±1.24 &  86.45±0.82   &  22.39±1.07    & 32.13±1.03 &32.26±1.10 &40.68±1.05\\
                     Pseudo Labeling$_{BERT}$ \cite{rosenberg2005semi}  &  56.56±1.07  &  66.38±1.02    &  80.42±0.58   & 83.38±0.69  & 60.37±1.09 & 68.70±1.61  & 84.26±0.66 &  86.79±0.58  & 22.52±1.30   &  32.28±0.92 &32.29±0.93 &40.99±0.96 \\
                    Noisy Student$_{BERT}$ \cite{xie2020self}   & \underline{56.67±1.11}   &  \underline{66.50±0.88}  &  80.78±0.47  &  \underline{83.68±0.92} & \underline{61.24±1.80}  & \underline{69.12±1.54}    & 84.87±0.70 &  \underline{87.21±0.67}  & 22.78±0.82  & 33.45±0.97 &33.64±1.07 &\underline{41.52±1.14} \\
                    Mixmatch$_{BERT}$ \cite{berthelot2019mixmatch}   &  56.52±0.80  & 66.34±1.22    & \underline{81.03±0.55}  & 83.14±0.98 & 61.18±0.92   & 69.05±1.01  & \underline{85.12±0.92} &86.94±0.72   & \underline{23.04±0.84}  & \underline{33.76±1.04} &\underline{33.72±1.29} & 41.27±1.08 \\
                     \textbf{{\modelname}$_{BERT}$ (Ours)}    &  \textbf{58.62±0.91}  &   \textbf{68.01±0.49}       &  \textbf{82.99±0.55}       &  \textbf{85.07±0.81}        & \textbf{63.60±0.77} & \textbf{70.65±0.59}  & \textbf{86.75±0.89}  &  \textbf{88.38±0.66}  &  \textbf{24.53±0.91}  &  \textbf{35.66±1.40}  &
                     \textbf{36.32±1.44}  &
                     \textbf{43.57±1.12}    \\\hline \hline
                     BERT w. gold labels   &  --  &    --  &   87.59±0.41       & 87.70±0.51        &  --  &  --  &  90.48±0.72  &    90.61±0.30    &  --  &  --   &  45.81±1.23  &  46.19±1.28    \\
\thickhline
\end{tabular}
}
\end{table*}

\begin{table*}[t]
\centering
\caption{F1 (\%) comparisons on the SemEval and TACRED datasets with various amounts of labeled data and 50\% unlabeled data for RE task.}
\label{tab:main_re}
  \resizebox{0.9\textwidth}{!}{
\begin{tabular}{lcccccccccc}
\thickhline
\multicolumn{1}{l}{\multirow{2}{*}{\textbf{Methods / \%Labeled Data}}} & \multicolumn{5}{c}{\textbf{SemEval}} & \multicolumn{5}{c}{\textbf{TACRED}} \\ \cmidrule(lr){2-6}\cmidrule(lr){7-11}
\multicolumn{1}{c}{}     & 5-shot     & 10-shot   & 5\%    & 10\%    & 30\%   & 5-shot   & 10-shot  & 3\%    & 10\%    & 15\%    \\ \hline
                          LSTM \cite{hochreiter1997long}     & 20.76±3.46   &  23.45±3.92   &  22.65±3.35        &   32.87±6.79       &   63.87±0.65  &  5.04±1.27    &  6.09±0.94   &  28.68±4.29       &  46.79±0.99       &   49.42±0.59    \\
                          PCNN \cite{zeng2015distant}       &   23.45±3.92 &   43.82±4.07    &  41.82±4.48       &    51.34±1.87     &  63.72±0.51       &  5.22±1.30  &6.73±1.35 &  40.02±5.23       &  50.35±3.28    &  52.50±0.39  \\
                           PRNN \cite{zhang2017position}     &  26.81±1.23  &  51.74±1.90    &  55.34±1.08    & 62.63±1.42  & 69.02±1.01 & 6.38±1.20   & 8.44±1.02  & 39.11±1.92  &  52.23±1.20  & 54.55±1.92   \\
                          BERT \cite{devlin2019bert}   &  32.09±0.78  &   59.62±0.81   &  70.71±1.24  &  71.93±0.99   &  78.55±0.87 &  10.23±1.11  & 14.70±1.19 & 40.11±3.88  & 53.17±1.67   &  55.55±0.82   \\\hline \hline
                       BiGAN$_{BERT}$ \cite{donahue2016adversarial}   &  34.26±1.52  & 61.87±1.44  & 72.38±1.44 & 73.94±1.21 & 80.46±1.00  & 11.46±0.88   & 16.08±0.94 &  42.31±1.36  &  54.78±1.48 & 56.35±1.03   \\
                        Mean Teacher$_{BERT}$ \cite{tarvainen2017mean}  & 35.83±0.96   & 62.58±1.13    & 73.29±1.63  & 74.86±0.84  & 82.90±1.03  & 12.44±0.93   & 16.83±0.88 & 43.21±1.88  &  55.63±1.39  & 56.77±0.83 \\
                     UDA$_{BERT}$ \cite{xie2020unsupervised}  &  35.92±1.03  & 63.04±0.90   & 73.82±1.56    & 75.38±0.77  & 82.88±0.86  & 12.50±1.02   & 16.94±0.92 & 43.56±1.69   & 55.55±1.21    & 56.92±0.81   \\
                      Deep Co-training$_{BERT}$ \cite{blum1998combining}     &  36.01±1.10  & 63.88±0.78 & 74.54±1.70    & 75.32±0.83  & 82.94±1.08 & 12.40±0.89   & 16.84±0.77 &  43.48±1.04   & 55.24±0.84    & 56.59±0.70 \\
                     Pseudo Labeling$_{BERT}$ \cite{rosenberg2005semi}  &  36.22±1.03  &  63.89±0.85  &  74.34±1.48  &  75.69±1.33   & 82.41±1.22  & 12.48±1.00   & 17.02±0.64 & 43.77±0.92   & 55.39±0.71    &  56.93±0.56   \\
                    Noisy Student$_{BERT}$ \cite{xie2020self}   & \underline{36.53±0.64}   & 64.66±0.73    &  74.68±0.83   & 76.11±1.06 &  \underline{82.95±0.46}  & 12.77±0.94   & \underline{17.11±0.75} &  \underline{44.06±0.89}  & 55.82±0.54  &  \underline{57.40±0.44}  \\
                    Mixmatch$_{BERT}$ \cite{berthelot2019mixmatch}   &  36.47±1.08  &  \underline{64.82±0.56}   &  \underline{75.26±0.92}   &  \underline{76.34±0.83}  & 82.86±0.50  & \underline{12.93±0.68}   & 17.05±0.93 & 43.92±0.83    & \underline{55.90±0.80}   &  57.28±0.82  \\
                     \textbf{{\modelname}$_{BERT}$ (Ours)}    &  \textbf{38.55±0.76}  &     \textbf{66.93±0.39}    &  \textbf{79.65±0.68}       &  \textbf{81.69±0.57}        & \textbf{85.52±0.34}  &  \textbf{15.59±0.50}  & \textbf{19.88±0.71}  &  \textbf{47.37±0.74}  &  \textbf{58.20±0.33}    &  \textbf{59.93±0.31}    \\\hline \hline
                     BERT w. gold labels   &  --  &   --   &   84.64±0.28       & 85.40±0.34        &   87.08±0.23  &  --  &  --  &    62.93±0.41    &  63.66±0.23        &   64.69±0.29     \\
\thickhline
\end{tabular}
}
\end{table*}
(1) \textbf{Generative methods}: \textbf{BiGAN} \cite{donahue2016adversarial} (Bidirectional Generative Adversarial Networks) is an unsupervised feature learning framework. BiGAN modifies the standard GAN structure by including an encoder that converts data $\boldsymbol{x}$ to $\boldsymbol{z^{'}}$, creating the pair of data $(\boldsymbol{x}, \boldsymbol{z^{'}})$. There are two types of true and fake data pairs: the pair $(\boldsymbol{x}, \boldsymbol{z^{'}})$ and the pair $(G(\boldsymbol{z}), \boldsymbol{z})$ produced by the generator $G$ with noise variables $\boldsymbol{z}$. The BiGAN discriminator $D$ needs to discriminate between true and fake data pairs in addition to classifying the data.

(2) \textbf{Consistency regularization methods}: \textbf{Mean Teacher} \cite{tarvainen2017mean} is jointly optimized by a perturbation-based loss and a training loss to ensure that the model makes consistent predictions on similar data. \textbf{UDA} \cite{xie2020unsupervised} (Unsupervised Data Augmentation) explores the function of noise injection in consistency training and replaces low-quality noise operations using back-translation \cite{sennrich2016improving} for text. The UDA extends the development in supervised data augmentation to semi-supervised learning by using the consistency regularization framework.

(3) \textbf{Graph-based methods}: Due to the difficulty of encoding sentence states with vanilla models and neighborhood aggregation operations with graph-based neural networks, we do not reproduce graph-based methods for comparison.
Yang \textit{et al.} \cite{yang2021survey} also clarified that graph-based methods do not perform well in deep low-resource information extraction methods; thus, losing comparisons with such methods does not affect our model in achieving state-of-the-art.

(4) \textbf{Self-training methods}: \textbf{Deep Co-training} \cite{blum1998combining} assumes that each data in the dataset has two distinct and complementary views and that each view is adequate for developing a strong classifier. The idea of co-training is to ensure that the predictions made by two classifiers on the same set of data are consistent.
\textbf{Pseudo Labeling} \cite{rosenberg2005semi} iteratively improves the model by predicting pseudo labels on unlabeled data and adds these pseudo label data to labeled data. 
\textbf{Noisy Student} \cite{xie2020self} suggests a semi-supervised strategy that is motivated by knowledge distillation. To create pseudo labels for unlabeled data, the teacher model is first trained on labeled data. Then, a larger model as a student is trained on both labeled and pseudo-labeled data. During training, the student model additionally includes dropout and stochastic depth.

(5) \textbf{Hybrid methods}: \textbf{Mixmatch} \cite{berthelot2019mixmatch} mixes consistency regularization and entropy minimization in a unified loss function. Mixmatch applies $K$ ($K=2$) times stochastic data augmentation to an unlabeled data, and each augmented data is fed through the classifier. Then, the temperature of the distribution is altered, ``sharpening'' the average of these $K$ ($K=2$) predictions.


Finally, we present another model: \textbf{BERT w. gold labels}, which indicates the upper bound of low-resource information extraction vanilla models when all unlabeled data has gold labels during training with labeled data.

For the evaluation metrics, we choose F1 score as the main metric. In the named entity recognition task, we follow a span-level evaluation setting where the named entity is considered correct when its boundary and category are both predicted correctly. In relation extraction task, following previous works \cite{hu2020semi}, the correct predictions of \textit{no\_relation} are ignored. In event extraction task, we choose Trig-C (Trigger Classification) and Arg-C (Argument Classification) for evaluation. Trig-C represents both the offsets and event type of a trigger match a reference trigger. Arg-C represents that the offsets, role and associated event type of an argument match a reference argument mention.

\subsection{Implementation Details}\label{implementation details} 
\subsubsection{Named Entity Recognition}
In our experiments, all training sets from three named entity recognition datasets are randomly sampled and divided into labeled and unlabeled datasets. We sample 5-shot, 10-shot, 5\%, and 10\% of the training set for OntoNotes 5.0 and CoNLL-2003 as labeled datasets. For WNUT 2017, we sample 5-shot, 10-shot, 3\%, and 10\% of the training set as labeled datasets. We sample 50\% of the training set of all three datasets as unlabeled datasets. Following previous works \cite{lin2019learning,hu2020semi}, we split the unlabeled dataset into 10 segments for all three datasets, and the model is optimized on one segment of the data in each iteration. \footnote{We give the implementation details of the vanilla models for the three tasks of NER, RE, and EE in the Appendix.}

\subsubsection{Relation Extraction}
For the two RE datasets, strictly following previous works \cite{lin2019learning,hu2020semi}, we use stratified sampling to divide the training set into labeled and unlabeled datasets of various proportions to ensure that all subsets share the same relation label distribution. For SemEval, we sample 5-shot, 10-shot, 5\%, 10\%, and 30\% of the training set, for TACRED, we sample 5-shot, 10-shot, 3\%, 10\%, and 15\% of the training set as labeled datasets. For both datasets, we sample 50\% of the training set as unlabeled dataset. As suggested in Li \textit{et al.} \cite{li2020exploit}, we split all unlabeled data into 10 segments. In each iteration, vanilla model is optimized based on one segment of the data. 

\subsubsection{Event Extraction}
Similar to the settings of NER and RE tasks described above, in event extraction task, we sample 5-shot, 10-shot, 3\%, 10\%, and 15\% of the training set for each dataset. The proportion of unlabeled dataset sampled from training set is 50\% and is split into 10 segments as well.

\subsection{Main Results}
Tables \ref{tab:main_ner}, \ref{tab:main_re}, \ref{tab:main_ee}, and \ref{tab:main_ee_2} show the mean and standard deviation F1 results with \textbf{5 runs} of training and testing on three IE sub-tasks when leveraging various labeled data and 50\% unlabeled data. All deep low-resource methods could gain performance improvements from the unlabeled data when compared with the vanilla model that only uses labeled data (BERT), which demonstrates the effectiveness of unlabeled data in the low-resource IE setting. We could observe that {\modelname} outperforms all baseline models consistently. More specifically, we give the analysis on the three IE subtasks respectively as follows: 

For NER task, compared with the previous SOTA model: Noisy Student, {\modelname} on average achieves 1.76\% higher F1 in OntoNotes 5.0, 1.71\% higher F1 in CoNLL-2003, and 1.89\% higher F1 in WNUT 2017 across various labeled data. When considering the standard deviation, {\modelname} is almost always the most robust model compared to the baselines. 

An interesting conclusion is that when labeled data is very scarce, e.g. 5-shot for OntoNotes 5.0, CoNLL-2003, and WNUT 2017, although each entity category has only 5 training samples, {\modelname} could achieve an average 2.02\% F1 boost compared with Noisy Student. When more labeled data are available, 10\% for three datasets, the average F1 improvement is consistent, but reduced to 1.53\%. We attribute the consistent improvement of our method to the explicit feedback mechanism that {\modelname} adopted via trial and error: we use Gradient Imitation as an alternative for the classification loss in optimizing vanilla models.
The guidance from the gradient direction, as a part of the gradient imitation process, is more instructive, explicit, and generalizable than the implicit signals from training directly on labeled data using the classification loss. This appealing character becomes more indispensable to the model performance in the learning process, especially when less labeled data are available.

This conclusion is also demonstrated in the EE task shown in Tables \ref{tab:main_ee} and \ref{tab:main_ee_2} on the event Trigger and Argument Classification, respectively. Compared to the previous SOTA model: Noisy Student, {\modelname} on average achieves 2.01\% higher Trigger Classification F1 and 2.83\% higher Argument Classification F1 on ACE05-E, 2.58\% higher Trigger classification F1 and 2.80\% higher Argument Classification F1 on ACE05-E$^{+}$ across various labeled data. When labeled data are very scarce, e.g. 5-shot for ACE05-E and ACE05-E$^{+}$, the average F1 improvement brought by {\modelname} can reach 2.70\% on both datasets.

For RE task, from Table \ref{tab:main_re}, we can observe that {\modelname} achieves good performance improvements on various ratios of labeled data compared to the previous SOTA model: Mixmatch. For example, {\modelname} could achieve 2.08\% F1 and 2.66\% F1 boost on 5-shot and 30\% SemEval labeled data compared with Mixmatch, respectively. For TACRED, {\modelname} could achieve 2.66\% F1 and 2.65\% F1 improvement on 5-shot and 30\% labeled data. We attribute the significant improvement to that the RE has a lower task difficulty than NER and EE, so {\modelname} can quickly force pseudo-labeled data to imitate the standard gradient descent direction under different amounts of labeled data.

\begin{table*}[t]
\centering
\caption{Trig-C F1 (\%) comparisons on the ACE05-E and ACE05-E$^{+}$ datasets with various amounts of labeled data and 50\% unlabeled data for EE task.}
\label{tab:main_ee}
  \resizebox{0.9\textwidth}{!}{
\begin{tabular}{lcccccccccc}
\thickhline
\multicolumn{1}{l}{\multirow{2}{*}{\textbf{Methods / \%Labeled Data}}} & \multicolumn{5}{c}{\textbf{ACE05-E}} & \multicolumn{5}{c}{\textbf{ACE05-E$^{+}$}} \\ \cmidrule(lr){2-6}\cmidrule(lr){7-11}
\multicolumn{1}{c}{}     & 5-shot     & 10-shot   & 3\%    & 10\%   & 15\% & 5-shot   & 10-shot  & 3\%    & 10\%  & 15\% \\ \hline
                          LSTM \cite{hochreiter1997long}     & 19.45±2.38   &   28.58±2.69    &  38.24±2.62       &  42.39±2.45  &  46.23±2.38   & 16.11±3.17    &  29.33±2.74     &  34.27±2.50       &  42.34±2.71 & 46.39±2.54  \\
                          PCNN \cite{zeng2015distant}       & 23.54±2.66   &  32.46±3.04      &   44.53±2.77      &  47.28±2.18   &  51.44±2.06 &    20.56±2.41     &   31.92±2.60      &  40.53±2.23  & 46.84±1.95  & 50.82±2.47 \\
                           PRNN \cite{zhang2017position}     &  27.48±2.74  &  38.44±2.28      & 49.77±2.31     & 53.46±1.96 &56.17±2.06  & 24.17±2.04 & 38.04±1.74    & 45.62±1.64  & 52.91±1.84 & 55.76±1.68 \\
                          BERT \cite{devlin2019bert}   &  36.21±1.08  &   47.00±1.10   &  58.85±1.61  &  61.91±1.68   & 64.75±0.93 & 34.88±1.24  &  46.87±0.77    & 53.67±0.99   &  61.29±1.04  & 63.51±0.84 \\\hline \hline
                       BiGAN$_{BERT}$ \cite{donahue2016adversarial}   &  38.33±1.73  & 47.68±1.24  & 59.24±1.22  & 62.36±1.21  &65.46±1.19 &  35.70±1.36  & 47.12±1.39    & 54.21±1.21  & 61.79±1.23  &64.02±1.17\\
                        Mean Teacher$_{BERT}$ \cite{tarvainen2017mean}  & 39.53±1.28   &  48.12±1.02   &  59.67±1.04  &62.76±1.34 & 65.82±1.04 &  35.69±1.08 &  47.35±0.90  &  54.70±0.93  & 61.84±0.81 &64.38±0.83\\
                     UDA$_{BERT}$ \cite{xie2020unsupervised}    & 39.64±1.08    & 48.34±0.88  & 59.78±1.07 &62.89±0.97 & 65.96±0.84   & 35.88±1.04  & 47.52±0.74   & 54.83±0.88     & 62.11±0.90 & 64.62±0.75  \\
                      Deep Co-training$_{BERT}$ \cite{blum1998combining}      &38.84±1.16     & 48.29±0.91  &59.74±1.11  & 62.83±1.04 &65.94±0.96  &35.66±0.90  & 47.29±1.22    & 55.23±1.02    &61.96±0.74  & 64.59±1.42  \\
                     Pseudo Labeling$_{BERT}$ \cite{rosenberg2005semi}     & 39.61±1.03    &  48.33±0.76   & 59.96±1.04  &62.94±0.93 & 66.23±0.82  & 35.92±0.62 &  47.66±0.93   &  55.34±0.76   & 62.22±0.63 & 64.78±0.58 \\
                    Noisy Student$_{BERT}$ \cite{xie2020self}      & \underline{39.83±0.94}    & \underline{48.47±0.63}  &  \underline{60.72±0.79}  & 63.07±0.65 & \underline{66.52±0.77} &36.13±0.71  & \underline{47.94±1.02}    & \underline{55.82±0.69}   & 62.49±0.63  &\underline{65.21±0.46} \\
                    Mixmatch$_{BERT}$ \cite{berthelot2019mixmatch}     &  39.74±0.84  & 48.51±0.79   & 60.56±0.66  &\underline{63.21±0.70} & 66.48±0.64  & \underline{36.20±0.44} & 47.90±0.53    &  55.71±0.82   &  \underline{62.57±0.71}  & 65.18±0.34\\
                     \textbf{{\modelname}$_{BERT}$ (Ours)}    &  \textbf{42.58±0.95}  &    \textbf{50.78±0.73}     &  \textbf{62.47±0.87}       &  \textbf{64.76±0.49}    &  \textbf{68.06±0.51}  & \textbf{38.30±0.29}  & \textbf{49.79±0.70}  &  \textbf{59.39±0.73}    &  \textbf{65.05±0.86} & \textbf{67.95±0.69}   \\\hline \hline
                     BERT w. gold labels       &     --     &    --     &   69.52±0.60  &  
                     69.79±0.69  & 69.91±0.90 & --  &   --   &  68.01±0.57        &   68.42±0.90  & 68.99±0.59  \\
\thickhline
\end{tabular}
}
\end{table*}

\begin{table*}[t]
\centering
\caption{Arg-C F1 (\%) comparisons on the ACE05-E and ACE05-E$^{+}$ datasets with various amounts of labeled data and 50\% unlabeled data for EE task.}
\label{tab:main_ee_2}
  \resizebox{0.9\textwidth}{!}{
\begin{tabular}{lcccccccccc}
\thickhline
\multicolumn{1}{l}{\multirow{2}{*}{\textbf{Methods / \%Labeled Data}}} & \multicolumn{5}{c}{\textbf{ACE05-E}} & \multicolumn{5}{c}{\textbf{ACE05-E$^{+}$}} \\ \cmidrule(lr){2-6}\cmidrule(lr){7-11}
\multicolumn{1}{c}{}     & 5-shot     & 10-shot   & 3\%    & 10\%   & 15\% & 5-shot   & 10-shot  & 3\%    & 10\%  & 15\% \\ \hline
                          LSTM \cite{hochreiter1997long}  &  2.79±0.51    & 7.27±0.98   &  8.66±1.23      &  17.24±1.06       & 22.18±1.17   &   2.81±0.45     & 6.84±1.20    & 9.21±1.35      &  17.18±1.09         & 24.35±1.24  \\
                          PCNN \cite{zeng2015distant}       &  4.12±0.79   &  8.81±1.13    &  10.72±1.04       &  21.53±1.14   & 26.55±1.02  &  3.06±0.79       &  7.46±1.08       &  11.05±1.66    & 21.20±1.11  &28.44±1.30 \\
                           PRNN \cite{zhang2017position}     &   5.03±1.17 & 11.06±0.92       &  14.09±1.15    & 27.69±0.97 &31.24±1.01 & 3.47±1.22 & 10.17±1.10    &14.26±1.98   &26.53±1.06  & 32.91±0.92 \\
                          BERT \cite{devlin2019bert}   &  8.59±0.63  &   15.10±0.81   &  19.78±1.03  &  33.42±1.17   & 36.19±0.95 & 7.66±1.10  &  14.86±0.89    & 19.39±1.82   &  32.64±1.33  & 38.03±1.51 \\\hline \hline
                       BiGAN$_{BERT}$ \cite{donahue2016adversarial}   & 9.62±1.04   & 16.24±0.74  & 21.07±1.11  & 34.17±1.04  & 37.24±1.17 & 8.34±1.35   &  16.04±1.07   & 21.76±1.70  & 33.11±1.07  &38.14±1.28\\
                        Mean Teacher$_{BERT}$ \cite{tarvainen2017mean}  & 9.83±0.83   &  16.84±0.79   &  21.34±1.20  &34.35±1.01 &37.59±0.80 & 8.47±0.92  &16.53±1.18    &  21.68±1.22  &33.90±1.00  &38.58±0.97\\
                     UDA$_{BERT}$ \cite{xie2020unsupervised}    &  10.43±0.74  &  16.92±0.85 & 21.46±1.16 &34.29±0.92 & 37.93±0.53   & 8.59±0.78 &16.87±0.83    &  21.94±1.15    & 33.99±0.81 & 38.66±1.04  \\
                      Deep Co-training$_{BERT}$ \cite{blum1998combining}      & 10.50±0.54    &  16.74±0.68 & 21.38±0.93 & 34.39±0.81 & 37.46±0.80 & 8.48±0.53 & 16.93±0.69    &  21.83±1.19   & 33.72±0.90 & 38.95±1.07  \\
                     Pseudo Labeling$_{BERT}$ \cite{rosenberg2005semi}     & 10.84±0.46    & 16.98±0.62    & 21.62±0.86  &34.66±0.59 & 37.59±0.60  & 8.42±0.56 & 16.85±0.71    &  22.61±1.28   &33.88±0.95  & 39.41±0.85 \\
                    Noisy Student$_{BERT}$ \cite{xie2020self}      & \underline{11.23±0.49}    & 17.23±0.55  & \underline{22.56±0.55}   & 34.92±0.77 & \underline{38.04±0.73} & \underline{9.43±0.57} & \underline{17.36±0.61}    &  \underline{23.38±1.11}  & \underline{34.25±0.83}  & \underline{40.01±0.56} \\
                    Mixmatch$_{BERT}$ \cite{berthelot2019mixmatch}     & 10.97±0.52   &  \underline{17.36±0.46}  & 22.53±0.87   &\underline{35.08±1.02} & 37.98±0.51  & 9.11±0.56 & 17.29±0.52    &   23.00±0.70  & 34.11±0.66   & 39.80±0.72\\
                     \textbf{{\modelname}$_{BERT}$ (Ours)}    &  \textbf{13.82±0.37}  &    \textbf{19.69±0.42}     &  \textbf{25.55±0.79}       &  \textbf{37.77±0.64}    &  \textbf{41.31±0.38}  & \textbf{12.62±0.71}  & \textbf{19.88±0.55}  &  \textbf{27.19±0.96}    &  \textbf{36.60±0.58} & \textbf{42.15±0.60}  \\\hline \hline
                     BERT w. gold labels       &     --     &    --     &   44.41±0.66  &  
                     45.33±0.92  & 46.02±0.72 & --  &   --   &  45.89±1.20        &   46.46±0.84  & 46.87±0.92  \\
\thickhline
\end{tabular}}
\end{table*}

\subsection{Analysis and Discussion}
\subsubsection{Effectiveness of unlabeled data}
\begin{figure}[bt!]
    \centering
    \includegraphics[width=0.94\linewidth]{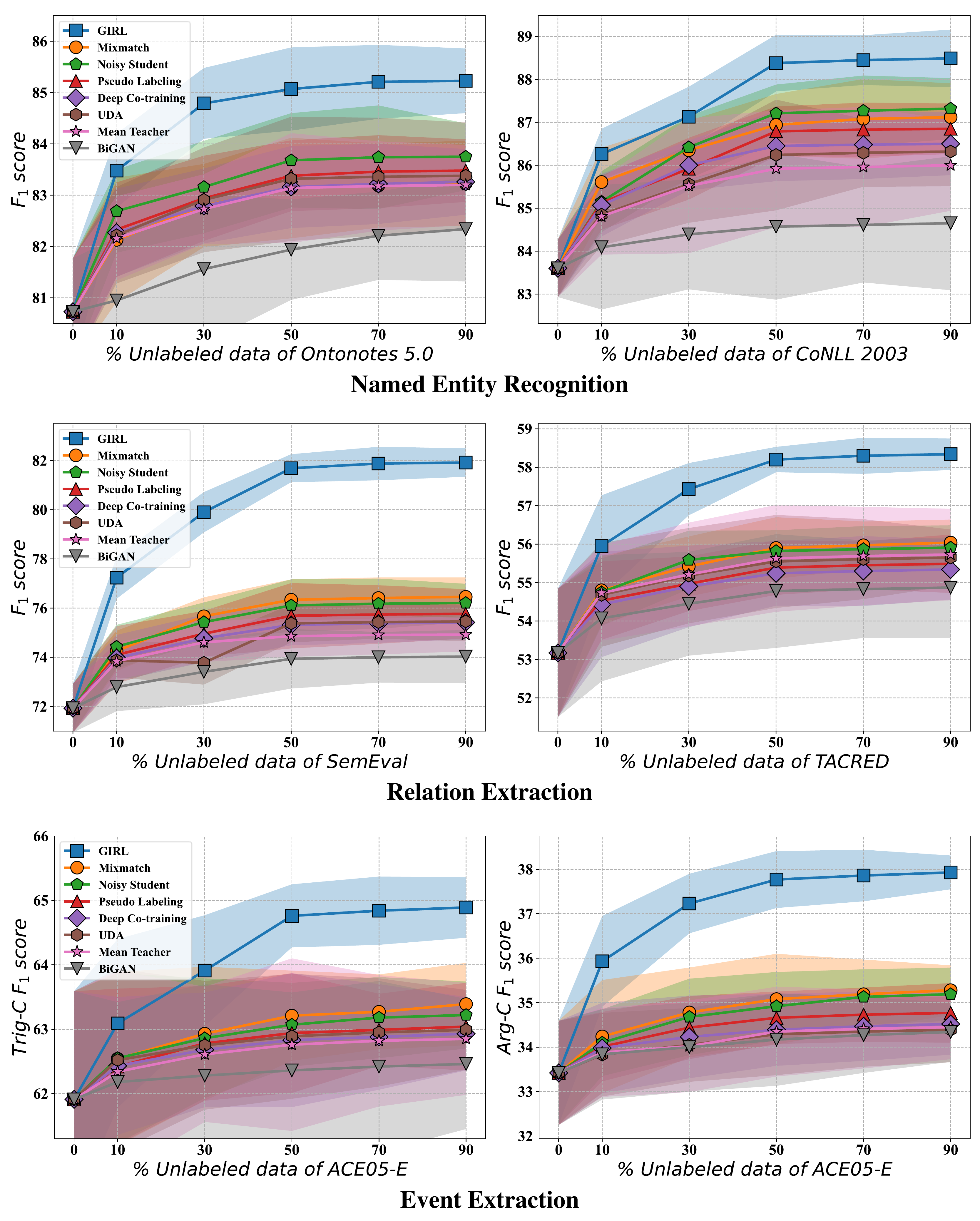}
    \caption{F1 (\%) Performance with various unlabeled data and 10\% labeled data on three IE sub-tasks.}
    \label{fig:unlabel}
\end{figure}

We further vary the ratio of unlabeled data on three IE sub-tasks and report performance in Figure \ref{fig:unlabel}. F1 performance on a fixed 10\% labeled data and 10\%, 30\%, 50\%, 70\%, and 90\% unlabeled data are reported. Note that both labeled data and unlabeled data come from the training set, so we can provide unlabeled data with an upper limit of 90\%. We could see that almost all methods have performance gains with the addition of unlabeled data and {\modelname} achieves consistently better F1 performance, with a clear margin, when comparing with baselines under all different ratios of unlabeled data. 

With the ratio of unlabeled data increases, we can observe that the magnitude of the improvement in F1 performance gradually decreases on the three IE sub-tasks, which is related to the common scheme of deep low-resource methods: the trained models will inevitably overfit on limited annotation, which impedes low-resource IE models from further generalizing beyond the limited annotations. Compared with other baseline models, {\modelname} obviously has better generalization ability since the F1 performance can still increase consistently with the increase of unlabeled data. Besides, {\modelname} has smaller variance on the three IE sub-tasks, which shows that {\modelname} is more robust than baseline methods.

\subsubsection{Ablation Study}
The main purpose of {\modelname} is to guide vanilla models to generate pseudo labels with the similar optimization directions as labeled data on the unlabeled data. {\modelname} minimizes the discrepancy between the gradient vectors obtained from the labeled data and the generated data. To demonstrate the effectiveness of the {\gradient} module, we conduct an ablation study where {\gradient} is removed from {\modelname} which is essentially the Pseudo Labeling$_{BERT}$ baseline. Pseudo Labeling$_{BERT}$ iteratively updates the model with the synthetic set containing labeled data and generated data without {\gradient}. From Tables \ref{tab:main_ner}, \ref{tab:main_re}, \ref{tab:main_ee}, and \ref{tab:main_ee_2}, we observe that {\modelname} w/o {\gradient} (Pseudo Labeling$_{BERT}$) gives us 2.44\%, 5.83\%, and 2.91\% loss on F1, averaged over all various amounts of labeled data on NER, RE, and EE tasks, respectively. 

\begin{figure}[t!]
    \centering
    \includegraphics[width=0.94\linewidth]{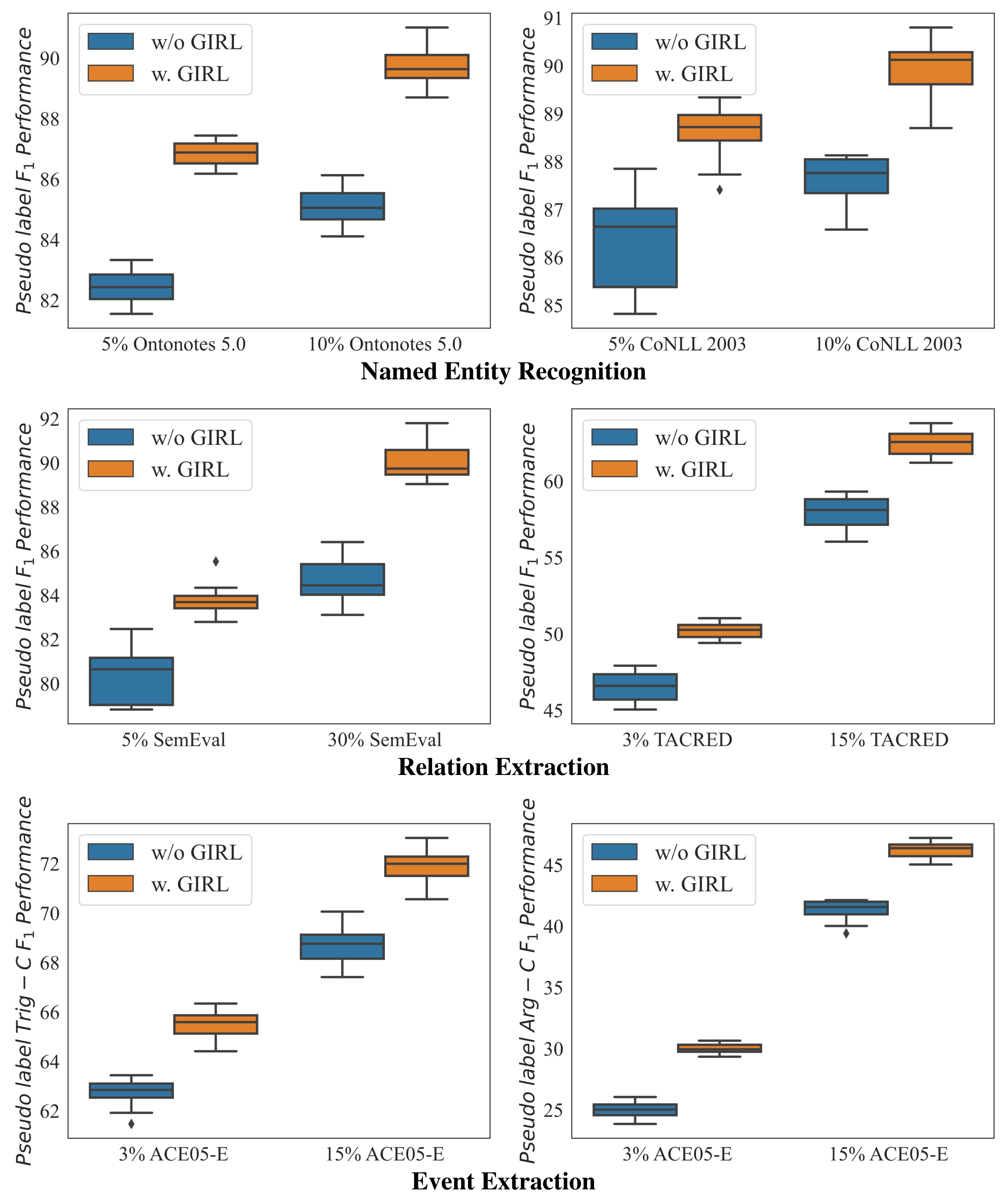} 
    \caption{Pseudo label F1 (\%) Performance with {\modelname} based on three IE sub-tasks.}
    \label{fig:PseudoF1}
\end{figure}

We argue that the performance gains of {\modelname} come from the improved pseudo label quality by adopting {\gradient}. 
To validate this, we draw a box plot to show the pseudo label F1. For the 10 unlabeled data segments used in 10 iterations, we report the F1 performance of pseudo labels by comparing with the golden labels on unlabeled data at each iteration, respectively. From Figure \ref{fig:PseudoF1}, we could find that for the three IE sub-tasks with different ratios of the labeled data, {\gradient} could undoubtedly improve the F1 performance of pseudo labels. In the case of 10\% CoNLL 2003, 30\% SemEval, 15\% TACRED, and 15\% ACE05-E where labeled data are less scarce, {\modelname} can obtain more accurate standard gradient directions based on an increased set of labeled data, compared to not using {\modelname}.
As a result, pseudo label performance improvements are more significant with {\gradient}. Considering the robustness of the pseudo label F1 performance on 10 iterations, the whiskers in the box plots after using {\modelname} are closer, which also reflects its stronger generalization ability.

\begin{figure}[t!]
    \centering
    \includegraphics[width=0.7\linewidth]{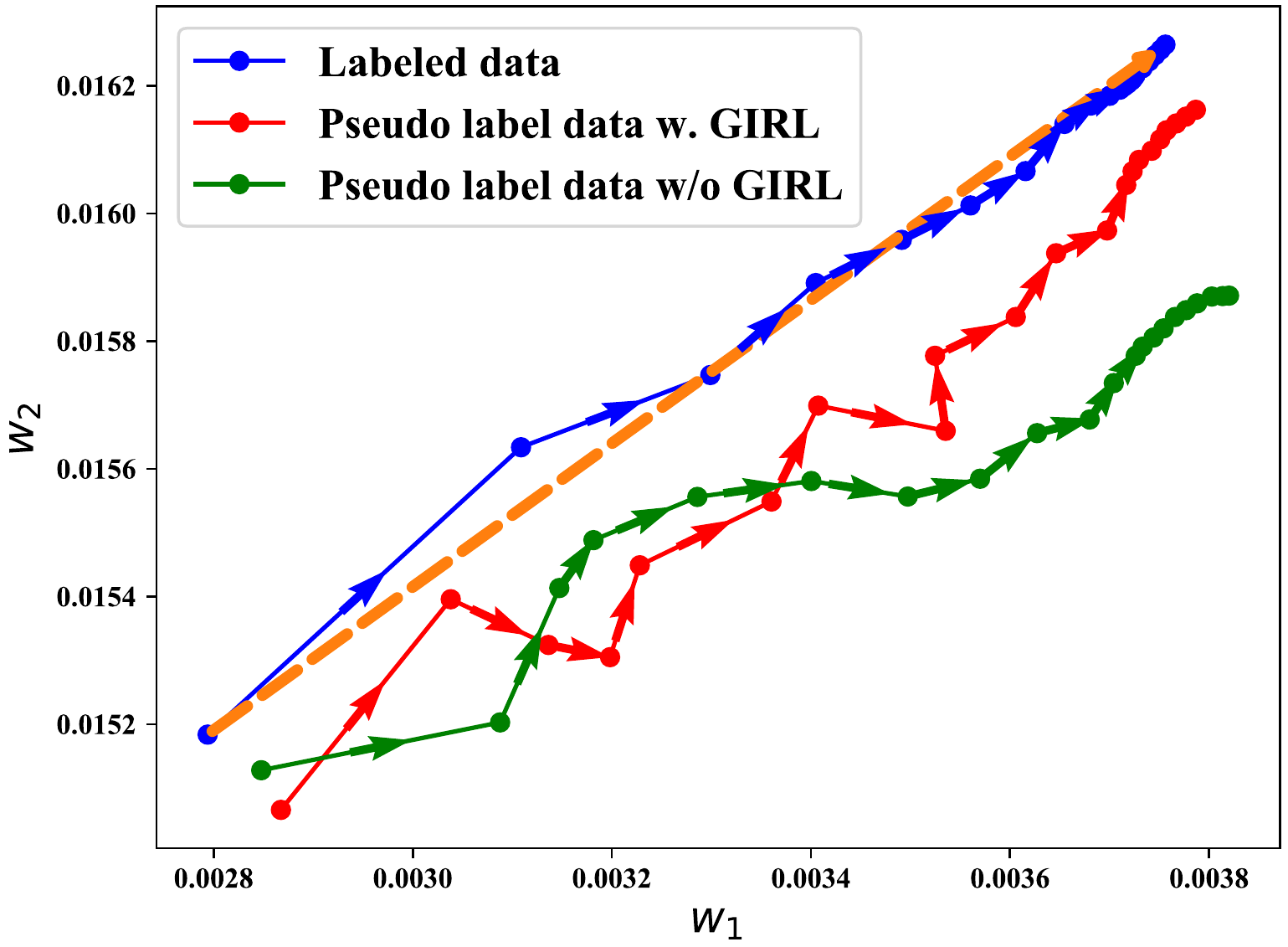}
    \caption{{\modelname} gradient descent directions on labeled data and pseudo label data. The dotted line indicates the average gradient direction on labeled data.}
    \label{fig:gradient}
\end{figure}

\begin{table*}[t!]
\centering
\caption{\textbf{Left:} Named Entity Recognition with/without {\modelname} on OntoNotes 5.0, where {\color{blue}\textit{blue}} represents the predicted named entity.\\\textbf{Right:} Relation Extraction with/without {\modelname} on SemEval, where {\color{red}\textit{red}} and {\color{blue}\textit{blue}} represent head and tail entities respectively.}
\label{tab:pseudo_label_case_study_ner+re}
\resizebox{0.85\textwidth}{!}{\begin{tabular}{@{}>{\bfseries\centering\arraybackslash}m{5em}|l|l}
\thickhline
``Other'' tag & \begin{tabular}[c]{@{}l@{}}{\color{blue}\textit{Jiujiang}} is the thoroughfare connecting Yangtze River's north...\\ {\color{purple}Label}: \textbf{\texttt{B-GPE}}\\{\color{purple}Prediction w/o GIRL}: \textbf{\texttt{O}}\\ {\color{purple}Prediction w. GIRL}: \textbf{\texttt{B-GPE}}\end{tabular} & \begin{tabular}[c]{@{}l@{}}My {\color{red}\textit{brother}} has entered my {\color{blue}\textit{room}} without knocking.\\ {\color{purple}Label}: \textbf{\texttt{Entity-Destination}}\\{\color{purple}Prediction w/o GIRL}: \textbf{\texttt{Other}}\\ {\color{purple}Prediction w. GIRL}: \textbf{\texttt{Entity-Destination}}\end{tabular}                                                                  \\
&&\\
Identify semantic nuances & \begin{tabular}[c]{@{}l@{}}Tanks currently are defined as armored vehicles weighing {\color{blue}\textit{25 tons}}\\ or more that carry large guns.\\ {\color{purple}Label}: \textbf{\texttt{B-QUANTITY, I-QUANTITY}}\\{\color{purple}Prediction w/o GIRL}: \textbf{\texttt{B-CARDINAL, O}}\\ {\color{purple}Prediction w. GIRL}: \textbf{\texttt{B-QUANTITY, I-QUANTITY}}\end{tabular} & \begin{tabular}[c]{@{}l@{}}The {\color{red}\textit{disc}} in a disc {\color{blue}\textit{music box}} plays this function, with pins\\ perpendicular to the plane surface...\\ {\color{purple}Label}: \textbf{\texttt{Content-Container}}\\{\color{purple}Prediction w/o GIRL}: \textbf{\texttt{Component-Whole}}\\ {\color{purple}Prediction w. GIRL}: \textbf{\texttt{Content-Container}}\end{tabular}\\
&&\\
Limitations & \begin{tabular}[c]{@{}l@{}}Defense lawyers have said all along that it was Palestinians and\\ not their Libyan clients who bombed {\color{blue}\textit{Pan Am 103}}.\\ {\color{purple}Label}: \textbf{\texttt{B-PRODUCT, I-PRODUCT, I-PRODUCT}}\\ {\color{purple}Prediction w/o GIRL}: \textbf{\texttt{O, O, O}}\\ {\color{purple}Prediction w. GIRL}: \textbf{\texttt{B-FAC, I-FAC, I-FAC}}\end{tabular} & \begin{tabular}[c]{@{}l@{}}{\color{red}\textit{Natural history}} {\color{blue}\textit{programmes}} began as live outside broadcasts\\ on BBC television in the early 1950s.\\ {\color{purple}Label}: \textbf{\texttt{Topic-Message}}\\ {\color{purple}Prediction w/o GIRL}: \textbf{\texttt{Other}}\\ {\color{purple}Prediction w. GIRL}: \textbf{\texttt{Other}}\end{tabular}                                            \\ \thickhline
\end{tabular}}
\end{table*}
\begin{table*}[t!]
\centering
\caption{\textbf{Left:} Trigger Identification and Classification in Event Extraction with/without {\modelname} on ACE05-E, where {\color{blue}\textit{blue}} represents the trigger.\\\textbf{Right:} Argument Identification and Classification in Event Extraction with/without {\modelname} on ACE05-E, where {\color{blue}\textit{blue}} and {\color{red}\textit{red}} represent the trigger and argument respectively.}
\label{tab:pseudo_label_case_study_ee}
\resizebox{0.85\textwidth}{!}{\begin{tabular}{@{}>{\bfseries\centering\arraybackslash}m{5.5em}|l|l}
\thickhline
``Other'' tag & \begin{tabular}[c]{@{}l@{}}The president of the United States coming for a brief two-day\\ {\color{blue}\textit{summit}} with the British prime minister Tony Blair.\\ {\color{purple}Label}: \textbf{\texttt{Contact:Meet}}\\{\color{purple}Prediction w/o GIRL}: \textbf{\texttt{No Identification}}\\ {\color{purple}Prediction w. GIRL}: \textbf{\texttt{Contact:Meet}}\end{tabular} & \begin{tabular}[c]{@{}l@{}}Reports suggesting airport {\color{red}\textit{buildings}} have been {\color{blue}\textit{attacked}}, but\\ runways remain operational.\\{\color{purple}Label}: \textbf{\texttt{Target-Arg}}\\{\color{purple}Prediction w/o GIRL}: \textbf{\texttt{No Identification}}\\ {\color{purple}Prediction w. GIRL}: \textbf{\texttt{Target-Arg}}\end{tabular}                                                                   \\
&&\\
Identify semantic nuances & \begin{tabular}[c]{@{}l@{}}The only people who profit from class {\color{blue}\textit{actions}} are the lawyers.\\ {\color{purple}Label}: \textbf{\texttt{Justice:Sue}}\\{\color{purple}Prediction w/o GIRL}: \textbf{\texttt{Justice:Appeal}}\\ {\color{purple}Prediction w. GIRL}: \textbf{\texttt{Justice:Sue}}\end{tabular} & \begin{tabular}[c]{@{}l@{}}At that point the baby's {\color{red}\textit{mother}} {\color{blue}\textit{stabbed}} him right in the hand.\\ {\color{purple}Label}: \textbf{\texttt{Agent-Arg}}\\{\color{purple}Prediction w/o GIRL}: \textbf{\texttt{Victim-Arg}}\\ {\color{purple}Prediction w. GIRL}: \textbf{\texttt{Agent-Arg}}\end{tabular}\\
&&\\
Limitations & \begin{tabular}[c]{@{}l@{}}U.S. troops thwarted a Baghdad bank robbery over the {\color{blue}\textit{protests}}\\ of Iraqis eager to share in the loot.\\ {\color{purple}Label}: \textbf{\texttt{Conflict:Demonstrate}}\\ {\color{purple}Prediction w/o GIRL}: \textbf{\texttt{No Identification}}\\ {\color{purple}Prediction w. GIRL}: \textbf{\texttt{No Identification}}\end{tabular} & \begin{tabular}[c]{@{}l@{}}If you were president, which national {\color{red}\textit{figures}} would you\\ {\color{blue}\textit{appoint}} to your cabinet and why?\\ {\color{purple}Label}: \textbf{\texttt{Person-Arg}}\\ {\color{purple}Prediction w/o GIRL}: \textbf{\texttt{No Identification}}\\ {\color{purple}Prediction w. GIRL}: \textbf{\texttt{No Identification}}\end{tabular}                                        \\ \thickhline
\end{tabular}}
\end{table*}

\subsubsection{Effectiveness of Gradient Imitation Reinforcement Learning}
We show the gradient descent direction of {\modelname} on labeled data and pseudo label data on the RE task in Figure \ref{fig:gradient}. Considering the overly-large parameters in the vanilla model, we use Principal Component Analysis \cite{wold1987principal} to reduce the dimension of the parameters to $2$, and reflect the direction of gradient descent according to the update of the parameters. Although the optimization direction of pseudo label data fluctuates at the beginning, {\modelname} is gradually improving and ends up closer to the ideal local minima. 


When {\gradient} is not used, the optimization direction is appealing at first because of the initial positive gains from the pseudo labeling schema. However, the error-prone pseudo labels obtained without instructive feedback gradually push the optimization away from the local minima, which leads to reduced generalization ability. For the vanilla models of NER and EE tasks, we get similar gradient descent direction figures by using the {\gradient} module, and we decide not to repeat the same gist here for conciseness.

\subsubsection{Case Study and Error Analysis}
We further study cases where pseudo labels are
improved with {\modelname} on three IE sub-tasks, and present them in Tables \ref{tab:pseudo_label_case_study_ner+re} and \ref{tab:pseudo_label_case_study_ee}. For NER, RE and EE tasks, a common phenomenon is that Prediction w/o {\modelname} tends to predict the pseudo label as \texttt{O}, \texttt{Other}, and \texttt{No Identification} with the most occurrences, most likely because these three labels being the dominating class in the dataset. Prediction w. {\modelname} is less sensitive to the label distribution in the data and assigns correct labels. 
More specifically, in the NER task of Table \ref{tab:pseudo_label_case_study_ner+re}, for the entity ``25 tons'', where ``25'' is a cardinal number and ``tons'' is a quantifier, the semantics of the two tokens are not similar, but {\modelname} is still able to associate it and give the correct \texttt{B-QUANTITY, I-QUANTITY} labels. Similarly, in the RE task, we could also observe cases where Prediction with {\modelname} is doing better at distinguishing the nuances between similar relations such as \texttt{Content-Container} and \texttt{Component-Whole} (Table \ref{tab:pseudo_label_case_study_ner+re}). For the EE task in Table \ref{tab:pseudo_label_case_study_ee}, in the Trigger Identification and Classification, although the Prediction w/o {\modelname} could identify the trigger word: ``actions'', it still has a hard time distinguishing between the classification categories \texttt{Justice:Sue} and \texttt{Justice:Appeal} which have similar semantics. In the Argument Identification and Classification, Prediction w/o {\modelname} is able to recognize that ``mother'' is an argument to the ``stabbed'' event, but mispredicts the argument type as \texttt{Victim-Arg} rather than \texttt{Agent-Arg}.

In the last case of Tables  \ref{tab:pseudo_label_case_study_ner+re} and \ref{tab:pseudo_label_case_study_ee}, we give error analysis for all IE sub-tasks. Prediction w. {\modelname} will predict incorrectly for samples with high difficulty or few occurrences. For example, in the NER task, ``Pan Am 103'' refers to the PA103 flight of Pan American Airlines, which is a proper noun, so the Prediction w. {\modelname} incorrectly classifies the entity type as \texttt{FACILITY} instead of \texttt{PRODUCT}. For the Argument Identification and Classification in the EE task, ``figures'' is an argument of \texttt{Person} in ``appoint'', but because ``figures'' is ambiguous and usually do not represent as person, the Prediction w. {\modelname} incorrectly classifies it as \texttt{No Identification}.
\section{Conclusions}\label{sec:conclusion}
In this paper, we propose a novel learning paradigm for low-resource Information Extraction. Different from conventional pseudo labeling models which endure gradual drift when generating pseudo labels, our reinforcement learning model {\modelname} encourages pseudo-labeled data to imitate the gradient optimization direction in the labeled data to improve the pseudo label quality --we find that our learning paradigm gives more instructive, explicit, and generalizable signals than the implicit signals that are obtained by training vanilla models directly with labeled data on IE sub-tasks. Experiments on seven public datasets across three IE sub-tasks (NER, RE, EE) in low-resource settings (semi-supervised IE and few-shot IE) show consistent improvements over competitive baselines.
\section{Limitations and Future Work}
In this section, we give {\modelname}'s weaknesses. {\modelname} adopts reinforcement learning to maximize the likelihood between the gradient optimization direction from pseudo-labeled data, and the standard gradient optimization direction on labeled data. Therefore, after each time step, it is necessary to calculate the Reward and update the State and Policy, which will lead to higher time complexity and computing resources. Therefore, a point that can be improved is how to reduce the time complexity of the algorithm by merging some similar States and Policies in the process of reinforcement learning, which is a feasible future research direction.
\ifCLASSOPTIONcompsoc



\ifCLASSOPTIONcaptionsoff
  \newpage
\fi



%


\normalem
\bibliographystyle{IEEEtran} 
\bibliography{custom}

%







\begin{IEEEbiography}[{\includegraphics [width=1in,height=1.25in,clip, keepaspectratio]{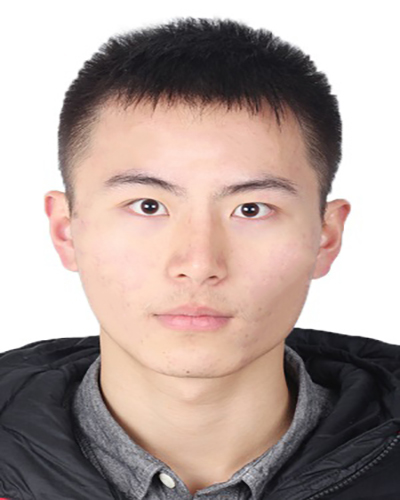}}] {Xuming Hu} received the B.E. degree in Computer Science and Technology, Dalian University of Technology. He is working towards the Ph.D. degree at Tsinghua University. His research interests include natural language processing and information extraction. \end{IEEEbiography}

\vspace{-10mm}

\begin{IEEEbiography}[{\includegraphics [width=1in,height=1.25in,clip, keepaspectratio]{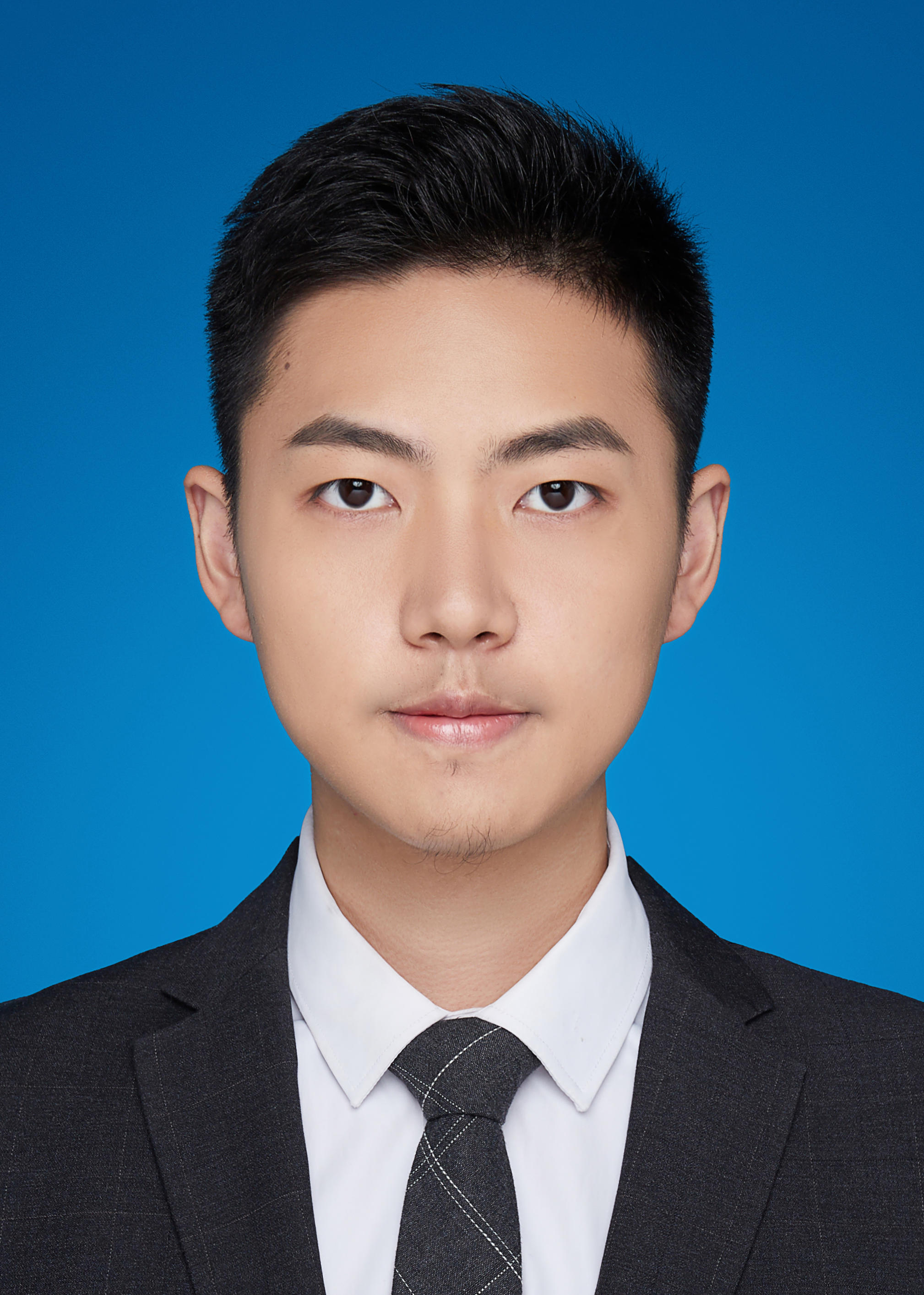}}] {Shiao Meng} received the B.E. degree in School of Software, Tsinghua University. He is working towards the Ph.D. degree at Tsinghua University. His research interests include natural language processing and information extraction. \end{IEEEbiography}

\begin{IEEEbiography}[{\includegraphics[width=1in,height=1.25in,clip,keepaspectratio]{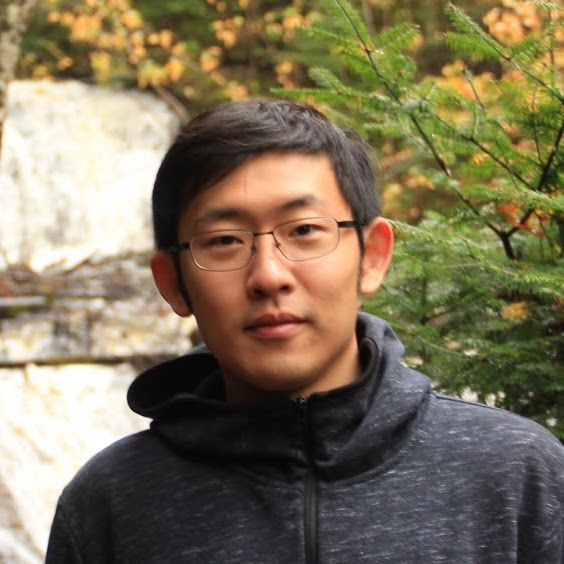}}]{Chenwei Zhang} is a senior applied scientist and tech lead at Amazon, Seattle, Washington. Before joining Amazon, he received the PhD degree in computer science from the University of Illinois at Chicago, Chicago, Illinois, in 2019. He is broadly interested in text/graph mining, natural language processing, and knowledge graphs. In particular, he is interested in text mining and mining structured information from heterogeneous information sources.
\end{IEEEbiography}

\vspace{-19mm}

\begin{IEEEbiography}[{\includegraphics[width=1in,height=1.25in,clip,keepaspectratio]{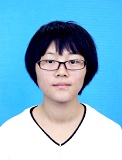}}]{Xiangli~Yang} is currently pursuing the Ph.D. degree with the School of Computer Science and Engineering, University of Electronic Science and Technology of China, Chengdu, China. Her research interests are semi-supervised learning, data mining, and machine learning.
\end{IEEEbiography}

\vspace{-19mm}

\begin{IEEEbiography}[{\includegraphics [width=1in,height=1.25in,clip, keepaspectratio]{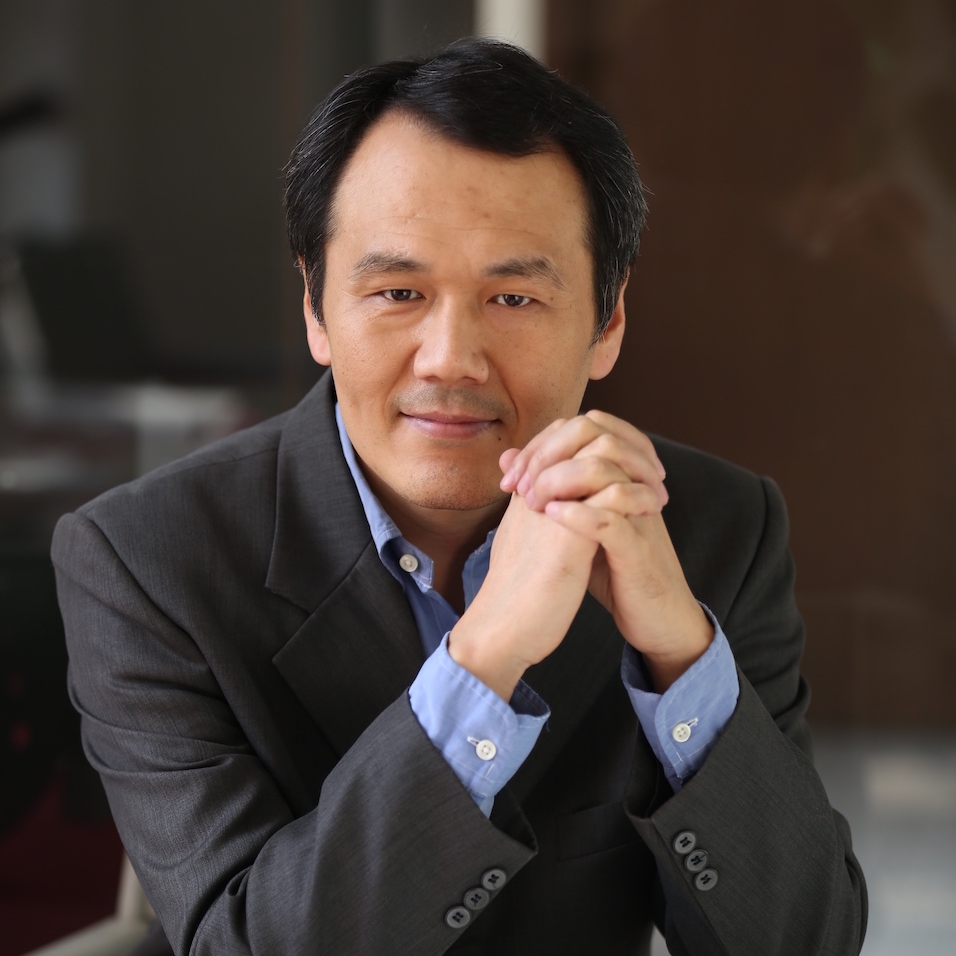}}] {Lijie Wen} received the B.S. degree, the M.S. degree, and the Ph.D. degree in Department of Computer Science and Technology, Tsinghua University, Beijing, China, in 2000, 2003, and 2007 respectively. He is currently an associate professor at School of Software, Tsinghua University. His research interests are focused on process data management, lifecycle management of computational workflow, and natural language processing. He has published more than 150 academic papers on conferences and journals, which are cited more than 4100 times by Google Scholar. \end{IEEEbiography}

\vspace{-14mm}

\begin{IEEEbiography}[{\includegraphics[width=1in,height=1.25in,clip,keepaspectratio]{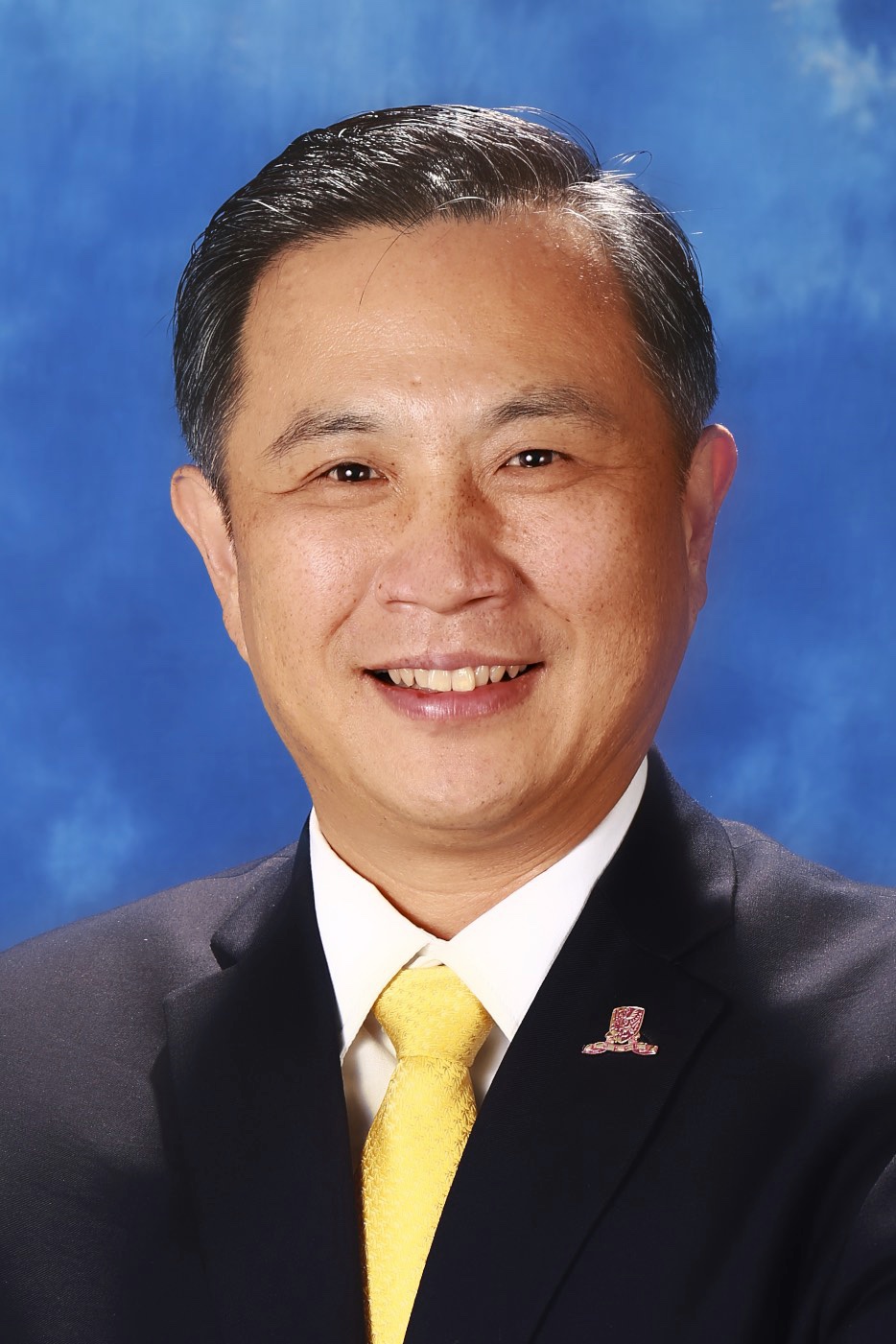}}]{Irwin~King (F’18)} is the Chair and Professor of Computer Science \& Engineering at The Chinese University of Hong Kong. His research interests include machine learning, social computing, AI, web intelligence, data mining, and multimedia information processing. In these research areas, he has over 300 technical publications in journals and conferences. He is an Associate Editor of the Journal of Neural Networks (NN). He is an IEEE Fellow, an ACM Distinguished Member, and a Fellow of Hong Kong Institute of Engineers (HKIE).  He has served as the President of the International Neural Network Society (INNS), General Co-chair of The WebConf 2020, ICONIP 2020, WSDM 2011, RecSys 2013, ACML 2015, and in various capacities in a number of top conferences and societies such as WWW, NIPS, ICML, IJCAI, AAAI, APNNS, etc. He is the recipient of the ACM CIKM 2019 Test of Time Award, the ACM SIGIR 2020 Test of Time Award, and 2020 APNNS Outstanding Achievement Award for his contributions made in social computing with machine learning. In early 2010 while on leave with AT\&T Labs Research, San Francisco, he taught classes as a Visiting Professor at UC Berkeley. He received his B.Sc. degree in Engineering and Applied Science from California Institute of Technology (Caltech), Pasadena and his M.Sc. and Ph.D. degree in Computer Science from the University of Southern California (USC), Los Angeles.
\end{IEEEbiography}

\vspace{-14mm}

\begin{IEEEbiography}[{\includegraphics [width=1in,height=1.25in,clip, keepaspectratio]{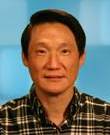}}] {Philip S. Yu} (Life Fellow, IEEE) is currently a Distinguished Professor and the Wexler Chair of information technology with the Department of Computer Science, University of Illinois Chicago (UIC), Chicago, IL, USA. Before joining UIC, he was with IBM Watson Research Center, where he built a world-renowned data mining and database department. He has authored or coauthored more than 780 papers in refereed journals and conferences. He holds or has applied for more than 250 U.S. Patents. His research interest include Big Data, including data mining, data stream, database, and privacy. He is a Fellow of ACM. Dr. Yu was the Editor-in-Chief of the ACM Transactions on Knowledge Discovery from Data during 2011–2017 and IEEE Transactions on Knowledge and Data Engineering during 2001–2004. He was the recipient of several IBM honors including the two IBM Outstanding Innovation Awards, Outstanding Technical Achievement Award, two Research Division Awards, and 94th Plateau of Invention Achievement Awards.
\end{IEEEbiography}

\clearpage

\appendix

\section*{More Implementation Details of Named Entity Recognition}
We use the pre-trained BERT to initialize vanilla model parameters and encode contextualized token-level representation. We use BERT-Base\_Uncased for WNUT 2017 and BERT-Base\_Cased for OntoNotes 5.0 and CoNLL-2003. We also use the corresponding BERT default tokenizer with max-length as 128 to preprocess the data. The tagging module above the encoder is a fully connected network with layer dimensions set as ${{h_{R}}}$-${h_{R}}$-label\_size, where ${h_{R}}=768$. We optimize our model using AdamW with learning rate and weight decay 3e-5 for all parameters (weight decay is the same as learning rate and both are selected from \{1e-5, 2e-5, 3e-5, 5e-5\}). The number of training epochs on the labeled data is selected from \{10, 20, 30, 40, 50\}. We adopt a linear warmup for the first 5\% steps. For \modelname, the total time step ${T}$ is set to 16, the same number as the batch size.

\section*{More Implementation Details of Relation Extraction}
For the vanilla model, we use the BERT default tokenizer with max-length as 128 to preprocess data. We use pretrained BERT-Base\_Cased as the initial parameter to encode contextualized entity-level representations. The fully connected network is defined with layer dimensions of ${2*{h_{R}}}$-${h_{R}}$-label\_size, where ${h_{R}}=768$. We use BertAdam with 1e{-4} (selected from \{5e-5, 1e-4, 2e-4\}) learning rate and warmup with 0.1 to optimize the loss. For {\modelname}, the time step ${T}$ is set to 16, the same number as the batch size. We use AdamW \cite{loshchilov2018fixing} with 5e{-5} (selected from \{1e-5, 5e-5, 1e-4\}) learning rate to optimize reinforcement learning loss.

\section*{More Implementation Details of Event Extraction}
We use BERT-Base\_Cased as the encoder and its corresponding default tokenizer to preprocess the data in both datasets. We optimize our model using AdamW with learning rate 2e-5 (selected from \{5e-6, 1e-5, 2e-5\}) for BERT and 2e-3 (selected from \{5e-4, 1e-3, 2e-3\}) for other parameters. Weight decay is set the same as learning rate. The number of training epochs in the labeled data is selected from \{10, 20, 30, 40, 50, 60\}. For the learning rate scheduler, we select the linear scheduler with a warmup ratio of 0.1. Following previous work \cite{lin2020joint}, we use the fully connected network with layer dimensions of ${{2*h_{R}}}$-${h}$-$label\_size$ for argument role classification, where $h_{R}=768$. For other subtasks, we use the layer dimensions of ${{h_{R}}}$-${h}$-$label\_size$. ${h=150}$ is used for entity extraction and ${h}=600$ is used for event extraction. The dropout rate for all classifiers is set to 0.4. For \modelname, the total time step ${T}$ is set to 10, the same number as the batch size.

\newpage

\section*{Summary of Difference}
Part of the study has been accepted as a long paper: ``Gradient Imitation Reinforcement Learning for Low Resource Relation Extraction'' in the 2021 Conference on Empirical Methods in Natural Language Processing (EMNLP 2021). We summarize the major extensions in this manuscript:
\begin{itemize}
\item We reconstructed the \texttt{GIRL} module to make it applicable to all IE sub-tasks, and demonstrated the effectiveness and generalization ability of \texttt{GIRL} in low-resource IE through experiments and analysis compared with strong baselines. More specifically, we revised \texttt{GIRL} in Section 4 (Proposed Framework) and added datasets for NER and EE in Section 5 (Experiments). We additionally compared two low-resource IE settings (semi-supervised IE and few-shot IE) and new generic baseline models (BiGAN, Mean Teacher, UDA, Deep Co-training, Pseudo Labeling, Noisy Student, and Mixmatch). Furthermore, a series of in-depth analyses were conducted to demonstrate the performance improvement of different modules in \texttt{GIRL} for general IE tasks. Through sufficient experiments, the effectiveness and robustness of \texttt{GIRL} are demonstrated.
\item We built vanilla models for three IE sub-tasks on which the \texttt{GIRL} can be adopted. More specifically, we gave the specific implementation process of the three vanilla models in Section 3 (Vanilla Models), and unified the framework description of the models as much as possible.
\item We revised the Abstract and Introduction Section to highlight how our method \texttt{GIRL} could handle general low-resource IE tasks.
\item We revised the Related Work Section to summarize deep low-resource methods and reinforcement learning methods in natural language processing.
\item We added the Limitations and Future Work Section to reveal the weaknesses of \texttt{GIRL}, and gave feasible future research directions.
\end{itemize}
\end{document}